\documentclass[10pt,journal,compsoc]{IEEEtran}

\ifCLASSOPTIONcompsoc
  \usepackage[nocompress]{cite}
\else
  \usepackage{cite}
\fi

\ifCLASSINFOpdf
\else
\fi

\usepackage{graphicx}
\usepackage{amssymb}  % $\blacktriangle$, $\bigstar$, and $\blacksquare$
\usepackage{bbm}  % \mathbbm
\usepackage{multirow}
\usepackage{booktabs}  % \toprule
\usepackage{bbding}  % \Solidbrush
\usepackage{amsmath,amsfonts}
\usepackage{url}
\begin{document}
\title{Collaborative Semantic Aggregation and Calibration for Federated Domain Generalization}

\author{Michael~Shell,~\IEEEmembership{Member,~IEEE,}
        John~Doe,~\IEEEmembership{Fellow,~OSA,}
        and~Jane~Doe,~\IEEEmembership{Life~Fellow,~IEEE}% <-this % stops a space
\IEEEcompsocitemizethanks{\IEEEcompsocthanksitem M. Shell was with the Department
of Electrical and Computer Engineering, Georgia Institute of Technology, Atlanta,
GA, 30332.\protect\\
E-mail: see http://www.michaelshell.org/contact.html
\IEEEcompsocthanksitem J. Doe and J. Doe are with Anonymous University.}% <-this % stops an unwanted space
\thanks{Manuscript received April 19, 2005; revised August 26, 2015.}}

\author{Junkun~Yuan,~\IEEEmembership{Student Member,~IEEE,}
        Xu~Ma,
        Defang~Chen,
        Fei~Wu,~\IEEEmembership{Senior Member,~IEEE,}
        Lanfen~Lin, ~\IEEEmembership{Member,~IEEE} and~Kun~Kuang*% <-this % stops a space
\IEEEcompsocitemizethanks{\IEEEcompsocthanksitem J. Yuan, X. Ma, D. Chen, L. Lin are with the College of Computer Science and Technology, Zhejiang University, Zhejiang, China.
K. Kuang is with Zhejiang University and Key Laboratory for Corneal Diseases Research of Zhejiang Province.
F. Wu is with Zhejiang University, Shanghai Institute for Advanced Study of Zhejiang University, and Shanghai AI Laboratory.
*Corresponding author. Junkun Yuan and Xu Ma contributed equally to this work. 
E-mail: \{yuanjk, maxu, defchern, kunkuang, llf\}@zju.edu.cn and wufei@cs.zju.edu.cn.}
\thanks{Manuscript received April 19, 2005; revised August 26, 2015.}}

\markboth{Journal of \LaTeX\ Class Files,~Vol.~14, No.~8, August~2015}%
{Shell \MakeLowercase{\textit{et al.}}: Bare Demo of IEEEtran.cls for Computer Society Journals}

\IEEEtitleabstractindextext{%
\begin{abstract}
Domain generalization (DG) aims to learn from multiple known source domains a model that can generalize well to unknown target domains. The existing DG methods usually exploit the fusion of shared multi-source data to train a generalizable model. However, tremendous data is distributed across lots of places nowadays that can not be shared due to privacy policies. In this paper, we tackle the problem of federated domain generalization where the source datasets can only be accessed and learned locally for privacy protection. We propose a novel framework called Collaborative Semantic Aggregation and Calibration (CSAC) to enable this challenging problem. To fully absorb multi-source semantic information while avoiding unsafe data fusion, we conduct data-free semantic aggregation by fusing the models trained on the separated domains layer-by-layer. To address the semantic dislocation problem caused by domain shift, we further design cross-layer semantic calibration with an attention mechanism to align each semantic level and enhance domain invariance. We unify multi-source semantic learning and alignment in a collaborative way by repeating the semantic aggregation and calibration alternately, keeping each dataset localized, and the data privacy is carefully protected. Extensive experiments show the significant performance of our method in addressing this challenging problem.
\end{abstract}

}

\maketitle

\IEEEdisplaynontitleabstractindextext
\IEEEpeerreviewmaketitle

\section{Introduction}\label{sec-int}
\IEEEPARstart{R}{ecently}, deep learning has made revolutionary advances to visual recognition \cite{he2016deep}, under the i.i.d. assumption that training and test data is sampled from the same distribution. Since the adopted datasets could be very distinct in many real-world applications, the performance of deep models learned from one training (source) dataset may drop rapidly on another test (target) dataset. To address this \emph{dataset/domain shift} \cite{quionero2009dataset} problem, \emph{domain generalization} (DG) \cite{blanchard2011generalizing, wang2022generalizing, zhou2021domains} is introduced to train a generalizable model to unknown target domains by learning from multiple semantically-relevant source domains. 

Numerous DG methods \cite{dou2019domain, Carlucci2019DomainGB, Zhao2020DomainGV} have been proposed recently. They popularize a variety of favorable strategies for training generalizable models by (indirectly) exploiting the fusion of \emph{``shared''} multi-source data. For example, some alignment-based methods \cite{li2018domain, Zhao2020DomainGV} match source data distributions in latent space for generating domain-invariant feature representations. Some meta-learning based strategies \cite{dou2019domain, Li2019EpisodicTF, li2019feature} utilize meta-train and meta-test datasets built by sampling from multi-source data for training a stable model to unknown domains. However, these methods may seriously violate data privacy policies, as tremendous data is stored locally in distributed places nowadays which may contain private information, e.g., the patient data from hospitals and the video recording from surveillance cameras. Therefore, a dilemma is encountered: The requirements of learning from shared multi-source data for training a highly generalizable model may hard to be met in many real scenarios due to the privacy issues. Meanwhile, without simultaneous access to the source datasets for obtaining adequate information of multi-source distribution, identifying and learning domain invariance for improving model generalization might be led astray. 

\begin{figure*}[t]
    \centering
    \includegraphics[trim={0cm 0cm 0cm 0cm},clip,width=1.3\columnwidth]{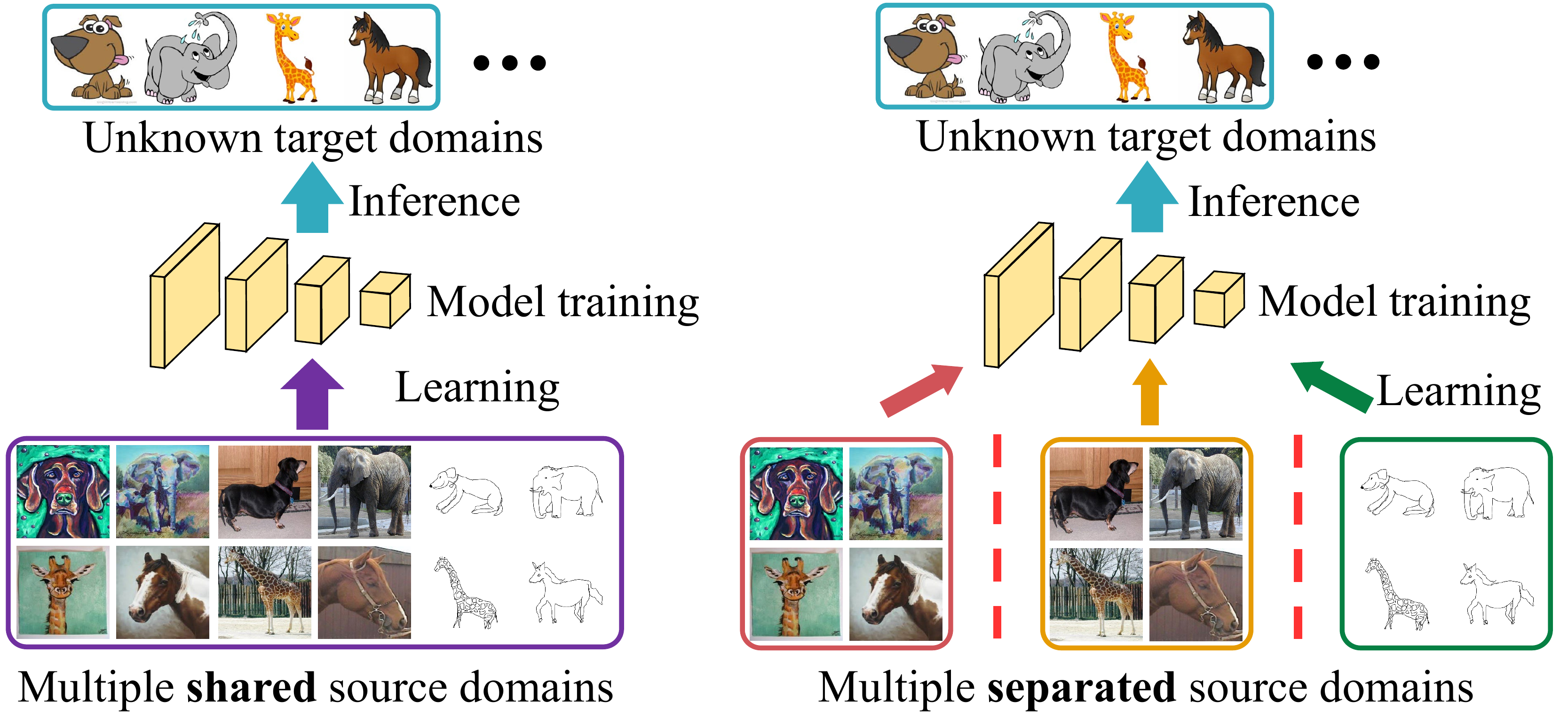}
    \caption{Comparison of the conventional domain generalization (left) and the federated domain generalization (right). The latter trains generalizable models while protecting privacy for many real-world scenarios.}
    \label{fig-sepdg}
\end{figure*}

In this paper, we tackle the problem of \emph{federated domain generalization} \cite{liu2021feddg} (see Fig. \ref{fig-sepdg}), where the source datasets are separated and can only be accessed and learned locally. It enables privacy preserving of sensitive data when employing them for improving model generalization. However, it is much more challenging than the conventional domain generalization task as: (1) The separated source datasets are private and may not be directly fused, hence the simultaneous learning of the multi-source semantic information is greatly hindered, making the identification of domain invariance tricky. (2) The heterogeneous source datasets with distinct data distributions may constitute enormous obstacles for training a generalizable model as the model is allowed to access only one local dataset each time, while the accessed dataset could contain particularly unusual bias and even bring negative gain for model generalization. 

We propose a novel method called Collaborative Semantic Aggregation and Calibration (CSAC) to enable federated domain generalization. We begin by hypothesizing that the deep models extract semantic information layer-by-layer, and the model parameters in each layer are related to the corresponding level as well as the training data distribution (proof-of-concept experiments are provided to verify it in Sec. \ref{sec-exp}). In light of this, to fully absorb multi-source semantic information while avoiding risky data fusion, a data-free semantic aggregation strategy is devised to fuse the models trained on the separated domains layer-by-layer. Then a semantic dislocation problem has arisen: Due to domain shift, the same level of semantic information from different domains could distribute across model layers during the aggregation. To this end, we further design cross-layer semantic calibration with an elaborate attention mechanism for precise semantic level alignment and domain-invariance enhancement. We unify the multi-source semantic learning and alignment in a collaborative way by repeating the semantic aggregation and calibration alternately. Each source dataset contributes semantic information locally for boosting model generalization during this process, resulting in a high-quality generalizable model under privacy protection. 

Our main contributions are summarized as: (1) We tackle a practical problem of federated domain generalization for addressing the dilemma between model generalization and privacy protection. This problem is important for training generalizable models in many privacy-sensitive scenarios but lacks extensive research to our knowledge.
(2) To enable federated domain generalization, we propose a novel framework called Collaborative Semantic Aggregation and Calibration (CSAC) to unify the multi-source semantic learning and alignment in a collaborative way by repeating semantic aggregation and calibration alternately.
(3) Extensive experiments on benchmark datasets show the significant performance of our method in addressing federated domain generalization, which is even comparable to the previous DG methods with shared source data. 

The rest of the paper is organized as follows. In Sec. \ref{sec-rel}, some related works about domain adaptation, domain generalization, federated learning, and distributed domain adaptation and generalization are introduced. In Sec. \ref{sec-met}, the problem definition of the federated domain generalization and our proposed CSAC framework and algorithm are stated. In Sec. \ref{sec-exp}, the results of the experiments on benchmark datasets as well as ablation studies and discussions are provided. We discuss the investigation of the federated domain generalization with a future outlook in Sec. \ref{sec-con}.

\section{Related Work}\label{sec-rel}
\subsection{Domain Adaptation}
To address the widespread domain shift problem, remarkable progress \cite{long2013adaptation, bitarafan2016incremental, pilanci2020domain, liu2018structure, jiang2018stacked, long2016deep, qiang2021robust, kuang2020causal, ren2020adversarial, li2021meta, ma2022attention, fan2020brain, yan2021transferable, dai2022graph, li2022dynamic, li2021faster, wu2020iterative, gao2023forget, wei2022dual, wei2018general} has been made in domain adaptation task. It aims to adapt a model trained on source domains to target domains by exploiting target data/information. 
One prevailing direction for this task is to employ adversarial learning \cite{qiang2021robust, li2021meta, dai2022graph} for reducing domain gap and generating domain-agnostic representations. 
Meanwhile, some algorithms \cite{long2013adaptation, long2016deep, li2022dynamic} are put forward to directly minimize domain divergence with distance metric like Maximum Mean Discrepancies (MMD). However, the target data/information is assumed to be available in this task, which greatly limits its implementation in many real-world applications since collecting adequate target data and information might be extremely expensive and laborious.

\subsection{Domain Generalization}
Domain generalization (DG) \cite{wang2022generalizing, zhou2021domains, blanchard2011generalizing, Zhao2020DomainGV, yuan2021domain, yuan2022label, Li2019EpisodicTF, niu2023knowledge, sun2022causal, miao2022domain, li2019feature, yuan2021learning, chen2022mix, ding2017deep} aims to train a stable model to unknown target domains by learning invariant knowledge from multiple source domains. A direct idea for DG is to align multi-source data distributions in latent space for generating invariant semantic representation \cite{li2018domain, li2018deep, Zhao2020DomainGV}. For example, Li et al. \cite{li2018deep} extract domain-invariant representations of multi-source joint distributions through a conditional invariant adversarial network.
Another set of works \cite{balaji2018metareg, dou2019domain, Li2019EpisodicTF, li2019feature} are based on meta-learning, they employ an episodic training paradigm that trains the model and improves its out-of-distribution generalization ability on meta-train and meta-test datasets, respectively, which are built by the shared multi-source data. For instance, Dou et al. \cite{dou2019domain} present a model-agnostic meta-learning training paradigm with two complementary losses to consider both global knowledge and local cohesion. 
Data augmentation \cite{Carlucci2019DomainGB, zhou2020deep, Wang2020LearningFE, zhou2021domain} for DG is also popular which trains the model on generated novel domains for improving model generalization. Among them, JiGen \cite{Carlucci2019DomainGB} is a representative work that utilizes the data with disordered patches to train the model  for solving a jigsaw puzzle. Some other works \cite{chattopadhyay2020learning, HuangWXH20} optimize the regularization terms of the data or networks to obtain generalization performance gain. 
These methods are mostly in thrall to shared multi-source data for identifying domain invariance and boosting model generalization, while concerns about data privacy are thus raised as tremendous private data might be distributed across separated places in many real scenarios. 
In comparison, we investigate a more practical setting of federated domain generalization towards privacy-preserving model training by accessing and learning each source dataset locally.

\subsection{Federated Learning} 
As an active research field towards modern privacy protection, \emph{federated learning} (FL) \cite{mcmahan2017communication, yang2019federated, zhang2020federated, lin2020ensemble, zhang2021unified, zhang2022towards, yoonfedmix, ghosh2020efficient, zhou2019privacy, li2021survey, xu2023data, huang2021feddsr, zhang2022robust} makes local clients jointly train a model with a central server and keeps data decentralized. 
Take a representative paradigm FedAvg \cite{mcmahan2017communication} as an example. In each communication round, a subset of the clients is chosen to receive the parameters of a global model from the server and trains it on their local data. The trained models are then transmitted back to the server for updating the global model with data-size based weights. Our investigated federated domain generalization task is closely related to federated learning as the data is decentralized, but the former is much more challenging: FL mainly focuses on guaranteeing model convergence when training on non-i.i.d. data \cite{yang2019federated}, and improving model performance on the \emph{``known''} clients. In contrast, our goal is to capture domain invariance from the separated source domains and train a generalizable model for the \emph{``unknown''} out-of-distribution target domains.

\begin{figure*}[t]
    \centering
    \includegraphics[trim={0cm 0cm 0cm 0cm},clip,width=1.98\columnwidth]{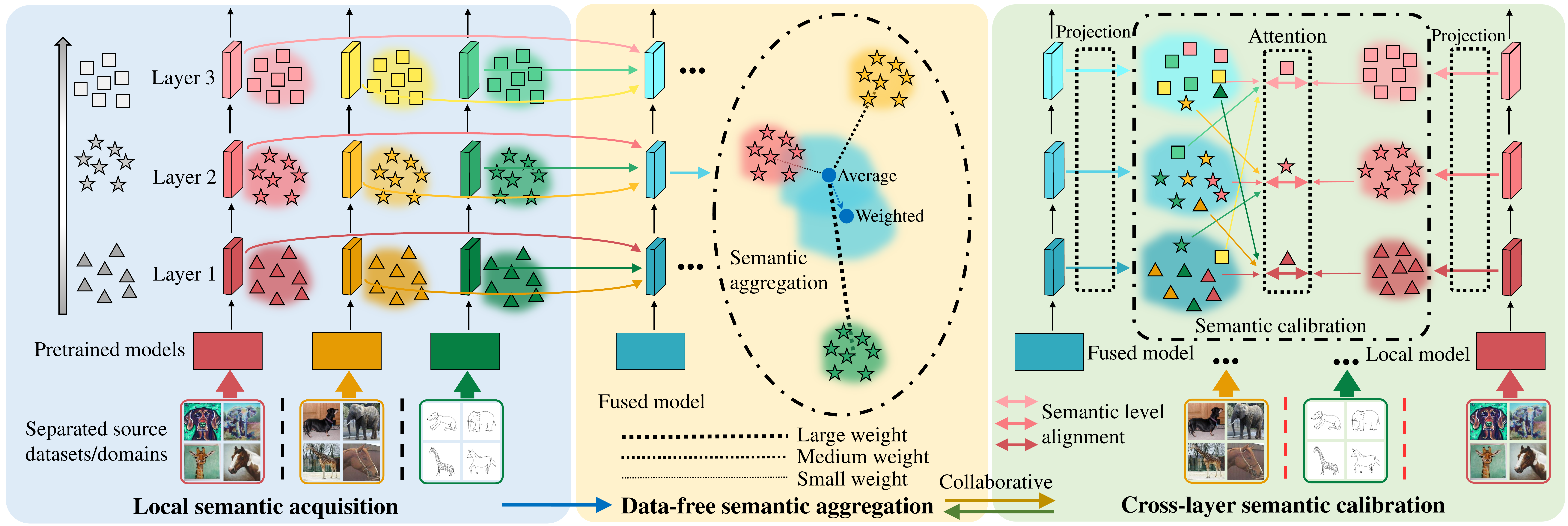}
    \caption{Overview of the proposed framework of Collaborative Semantic Aggregation and Calibration (CSAC). Left: To learn data distribution information of the local data, one model is trained on each source domain to extract semantics layer-by-layer, i.e., $\blacktriangle$, $\bigstar$, and $\blacksquare$. Middle: The trained models are fused layer-by-layer with semantic divergence based weights for data-free semantic gathering. Right: Cross-layer feature pairs between the fused model and a local model are matched with attention on each source domain for semantic level alignment and domain invariance enhancement. After semantic acquisition, the semantic aggregation and calibration processes are repeated alternately.
    }\label{fig-overview}
\end{figure*}

\subsection{Federated Domain Adaptation and Generalization}
Source-free domain adaptation \cite{kundu2020universal, li2020model} improves performance on the target domain by using a source pretrained model. Federated learning-based domain adaptation \cite{peng2019federated, jiang2023federated} adapts models from distributed source domains to target domains. 
However, these methods are limited to the strong assumption of available target data/information, as we discussed previously. 
To tackle this issue, lots of federated learning-based domain generalization methods \cite{chen2023federated, wu2021collaborative, zhang2021federated, nguyen2022fedsr, gupta2022fl, chen2022fraug, de2022mitigating, tian2021privacy, frikha2021towards, wei2022multi} have been proposed. 
\cite{nguyen2022fedsr} aims to learn simple representation of the distributed data with L2-norm and conditional mutual information constraints. 
FedDG \cite{liu2021feddg} is a representative method, which augments the distributed data in frequency space. However, it builds an amplitude spectrum distribution bank from the source data and shares it to all the clients, which might be time-consuming and needs high communication costs. Meanwhile, it shares the amplitude spectrum of the source data to all the clients that may increase the risks of data privacy leakage. 
In comparison, our proposed method do not have extra time-consuming procedures like building such a distribution bank. More importantly, we do not share any data (or parts of its information) during training for efficient communication and effective data privacy protection. 

Beyond computer vision, distributed training a generalizable model has also been explored in other fields, like natural language processing \cite{lin2021fednlp, guo2022federated, wang2022fedkc, weller2022pretrained, zhou2020progress, lit2022federated}, recommender system \cite{imran2023refrs, yang2020federated, wang2021fast}, speech recognition \cite{sattler2019robust}, edge computing \cite{tan2022towards}, etc. For example, \cite{wang2022fedkc} proposes a plug-and-play knowledge composition module to share knowledge across clients for non-i.i.d. multilingual natural language understanding. \cite{wang2021fast} proposes a recommendation model on decentralized domains, which learns data from user devices and trains a robust model by clipping training gradient. 
Compared with these works, we aim to solve the visual federated domain generalization problem by distributed learning invariant semantics of images through semantic aggregation and calibration processes for privacy protection.

\section{Method}\label{sec-met}
We begin with the problem definition of the federated domain generalization and its challenges for generalizable model learning. We then introduce our method CSAC (see Fig. \ref{fig-overview}) for addressing this challenging problem in detail. 

\subsection{Federated Domain Generalization}
In federated domain generalization, given source datasets $\{\mathcal{D}^{1}$, ..., $\mathcal{D}^{H}\}$ from $H$ distributed domains. There are $N^{h}$ data sampled from domain-specific distribution $P(X^h, Y^h)$ in each dataset $\mathcal{D}^h$, i.e., $\mathcal{D}^{h}=\{(x_{i}^{h},y_{i}^{h})\}_{i=1}^{N^{h}}$, defined on image and label spaces $\mathcal{X}\times{\mathcal{Y}}$. The goal is to utilize the distributed source datasets for training a generalizable model, which can perform well on unknown target domains. 

The challenges of this task are: 
(1) The source datasets are separated and can only be utilized locally, which greatly hinders the simultaneous learning of the multi-source semantic information and even leads to invalid domain invariance identification; 
(2) The heterogeneous source datasets with distinct data distributions constitute enormous obstacles for generalizable model learning, since the model can only access one local dataset each time. That is, if the exploited dataset contains unusual domain-specific bias, the trained model may even exhibit a negative generalization gain.

\subsection{Overview of Our Method}
The key idea of our method CSAC (Fig. \ref{fig-overview}) is to fully absorb multi-source information and precisely align semantic levels, which contains three main processes: (1) \emph{Local semantic acquisition} for learning distribution information of local data. (2) \emph{Data-free semantic aggregation} for semantic information gathering from the trained models. (3) \emph{Cross-layer semantic calibration} for semantic level alignment and domain invariance enhancement. 
After obtaining local distribution information in process (1), the latter two processes are repeated alternately, unifying the multi-source semantic learning and alignment in a collaborative way for generalizable model training. 
Note that by following the algorithms of federated learning \cite{mcmahan2017communication, yang2019federated, lin2020ensemble, yoonfedmix, ghosh2020efficient}, we only transmit models among the distributed domains, and neither data nor its information is shared, which adequately preserves privacy. 
To our knowledge, the practice of using the same model structure for heterogeneous data is widely adopted in domain generalization and federated learning researches, like \cite{yoonfedmix, liu2021feddg}. Therefore, we argue that it is feasible for our method to adopt the same model architecture for heterogeneous source data due to the powerful representation learning ability of deep models, as demonstrated by both the previous works and our experiments.

\subsection{Local Semantic Acquisition}
Before gathering and aligning the multi-source semantic information, we need to fully obtain the data distribution information of the separated source datasets. To avoid unsafe data fusion, we assign one model on each of the separated domains to impose data distribution learning. 
Given $H$ separated source datasets, $y^{h}$ with $C$ categories is the groud-truth label of the image $x^{h}$ in dataset $\mathcal{D}^{h}$, where $h\in\{1,...,H\}$. 
Let $\{G^{h}\}_{h=1}^{H}$ be the trained models, each model $G^{h}$ can be optimized on the local source data $\mathcal{D}^{h}$ with the following cross-entropy loss:
\begin{equation}
    \mathcal{L}_{CE}^{h}=-\mathbb{E}_{(x^{h},y^{h})\in\mathcal{D}^{h}}[\sum_{c=1}^{C}\mathbbm{1}(y^{h}=c)\log{G^{h,c}(x^{h})}],
\end{equation}
where $G^{h,c}$ is the $c$-th dimension of the output of model $G^{h}$. $\mathbbm{1}(\cdot)$ is an indicator function that equals to $1$ for the correct condition and 0 for the rest. 
To facilitate the following semantic aggregation and calibration processes, we further introduce label smoothing \cite{Mller2019WhenDL, liang2020we} to encourage data representations to group in tight evenly clusters, preventing the trained models from being over-confident. The updated learning loss for each trained model $G^{h}$ is
\begin{equation}\label{equ:cels}
    \mathcal{L}_{LS}^{h}=-\mathbb{E}_{(x^{h},y^{h})\in\mathcal{D}^{h}}[\sum_{c=1}^{C}p^{h,c}\log{G^{h,c}(x^{h})}],
\end{equation}
where $p^{h,c}=(1-\alpha)\mathbbm{1}(y^{h}=c)+\alpha/C$ is the smoothed label. $\alpha$ is a smoothing hyper-parameter empirically set to $0.1$ \cite{Mller2019WhenDL}.

\subsection{Data-Free Semantic Aggregation}
After acquiring distribution information of local data, we devise a data-free semantic aggregation strategy with the trained models $\{G^{h}\}_{h=1}^{H}$. 
Inspired by recent researches \cite{Wang2018InterpretNN, Zhang2019InterpretingCV} on the interpretability of deep neural networks, we hypothesize that the deep models extract semantic information layer-by-layer, and the model parameters in each layer are related to the corresponding semantic level as well as the training data distribution (proof-of-concept experiments are shown in Sec. \ref{sec-exp}). To this end, we propose to fuse the trained models layer-by-layer for gathering different levels of semantics from the separated source domains. 
Since each model $G^{h}$ trained on the data $\mathcal{D}^{h}$, it extracts hierarchical semantics from distribution of $\mathcal{D}^{h}$. 
Let $G_{l}^{h}$ be the $l$-th layer of $G^{h}$, we have the average model parameter distribution in the $l$-th layer, that is,
\begin{equation}
    G_{l}^{AVG}=\frac{1}{H}\sum_{h=1}^{H}G^{h}_{l}.
\end{equation}
We find that if a source data have a distribution far from the others, the parameters of the model trained on it would be distinct from the parameters of the other models. Therefore, this model would be considered less as it is far from the average distribution $G_{l}^{AVG}$ if we directly use $G_{l}^{AVG}$ as the final model. 
To fairly fuse the information of the source datasets for precise semantic calibration, we assign weight to each model based on its semantic divergence to $G_{l}^{AVG}$:
\begin{equation}\label{equ:fus}
    M_{l}=\sum_{h=1}^{H}\frac{\mathrm{dist}(G_{l}^{h},G_{l}^{AVG})}{\sum_{h=1}^{H}\mathrm{dist}(G_{l}^{h},G_{l}^{AVG})}G_{l}^{h},
\end{equation}
where $M_{l}$ is the $l$-th layer of the fused model $M$, $\mathrm{dist}(\cdot,\cdot)$ is distance metric and we empirically use $L_{2}$ distance (more discussions about it are in Sec. \ref{sec-exp}). 
Models with distinct parameters, or training data distributions, will be paid more attention to by being given a large weight. 
As the trained models are fused layer-by-layer, different levels of semantics from the separated source domains are aggregated for domain invariance learning in the semantic calibration.

\subsection{Cross-Layer Semantic Calibration}
Due to the different data distributions of the source datasets, i.e., domain shift, the same level of semantic information from different domains could be distributed across the layers of the fused model $M$ during the aggregation process, which we call the \emph{semantic dislocation} problem. 
To calibrate each level of semantic information for improving model generalization, we align each cross-layer semantic feature pair between the fused model $M$ and a local model $L^{h}$ (trained on each local source domain like $G^{h}$). 
Take Fig. \ref{fig-overview} (right) as an example. The second layer of each local model mainly contains the second level of semantic information, i.e., $\bigstar$. We match it with each layer of the fused model to align the second semantic level, i.e., match $\bigstar$ in different layers of the fused model, where each cross-layer pair is weighted by their semantic similarities. 
Since the hierarchical semantic features may have different size, we first project them to the same size (we use the size of the semantic features in the last adopted layer in the experiments):
\begin{equation}
\begin{aligned}
    M_{l}(f_{l})'=&Proj(M_{l}(f_{l})), \\ L_{m}^{h}(f_{m}^{h})'=&Proj(L_{m}^{h}(f_{m}^{h})),
\end{aligned}
\end{equation}
where $Proj(\cdot)$ is the projection function with one convolution layer (see experiments for details), $f_{l}$ and $f_{m}^{h}$ are the input features for the $l$-th of the fused model and the $m$-th layer of the local model, respectively. We align each cross-layer semantic feature pair $(l,m)$ after projection:
\begin{equation}
    \mathcal{L}_{AL}^{h}=\sum_{l\in\mathcal{R}}\sum_{m\in\mathcal{R}}\alpha_{l,m}D(M_{l}(f_{l})',L_{m}^{h}(f_{m}^{h})'),
\end{equation}
where $D(\cdot,\cdot)$ is used to measure the distribution discrepancy. We minimize $\mathcal{L}_{AL}^{h}$ to optimize the fused model $M$ on each domain $h$ for performing semantic feature alignment. We adopt Maximum Mean Discrepancies (MMD) \cite{gretton2012kernel} for $D(\cdot,\cdot)$ by following \cite{Long2016UnsupervisedDA, venkateswara2017deep}. 
The set of layers $\mathcal{R}$ for alignment is given in experiments. A dynamic weight $\alpha_{l,m}$ for the layer pair $(l,m)$ is based on the semantic similarity learned by the attention mechanism introduced in the following. 

\begin{figure*}[t]
    \centering
    \includegraphics[trim={0cm 0cm 0cm 0cm},clip,width=0.6\columnwidth]{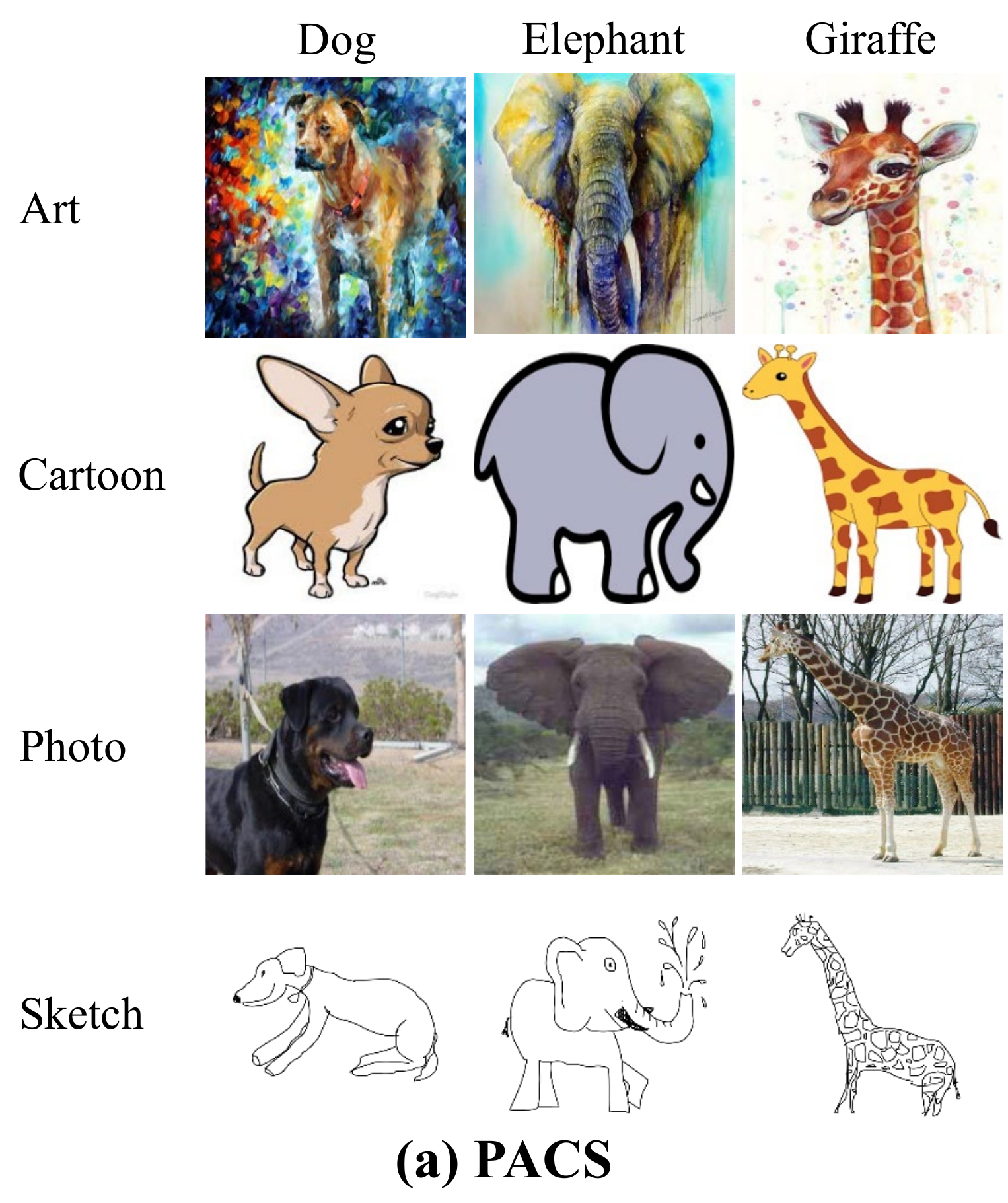}
    \includegraphics[trim={0cm 0cm 0cm 0cm},clip,width=0.6\columnwidth]{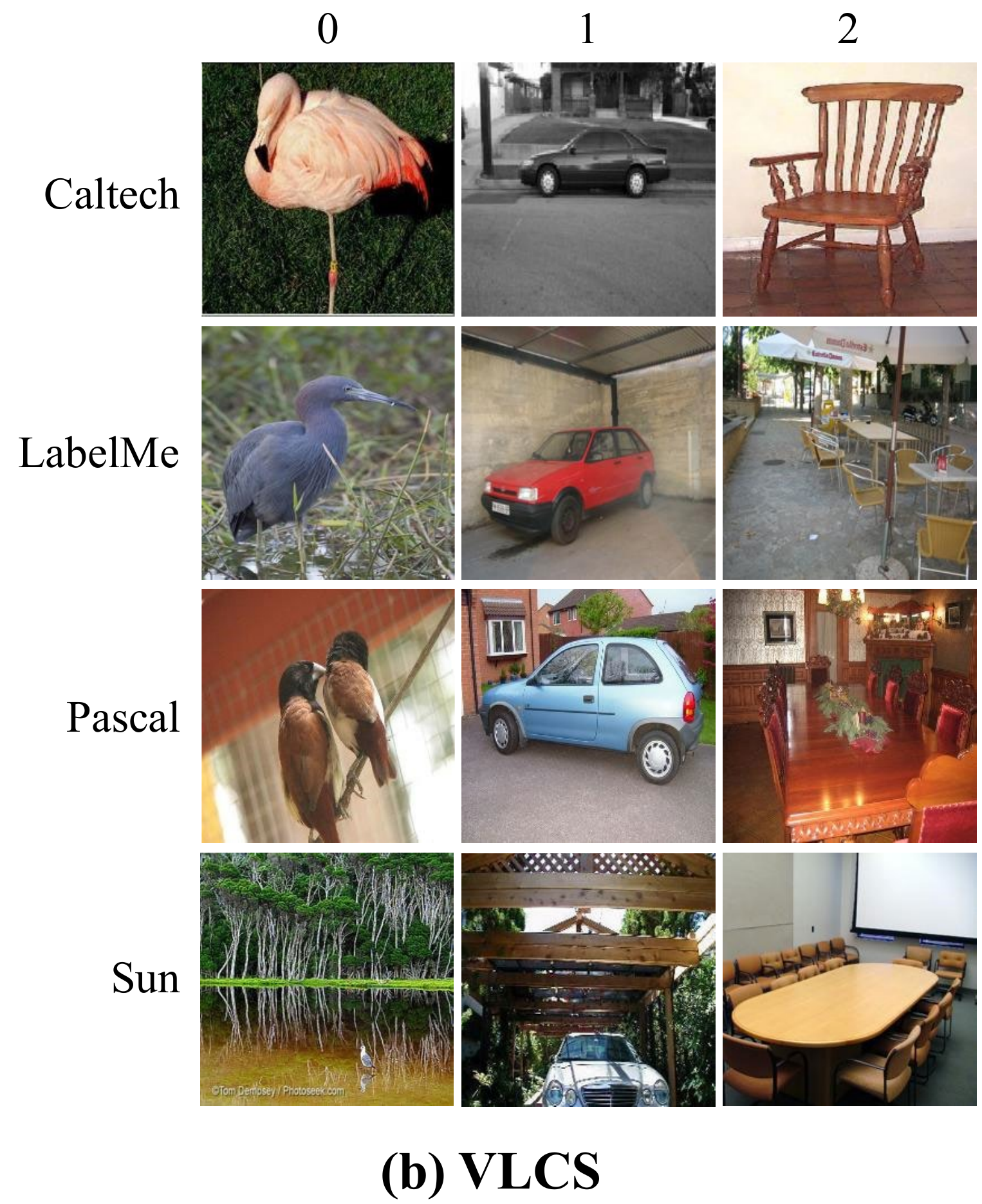}
    \includegraphics[trim={0cm 0cm 0cm 0cm},clip,width=0.8\columnwidth]{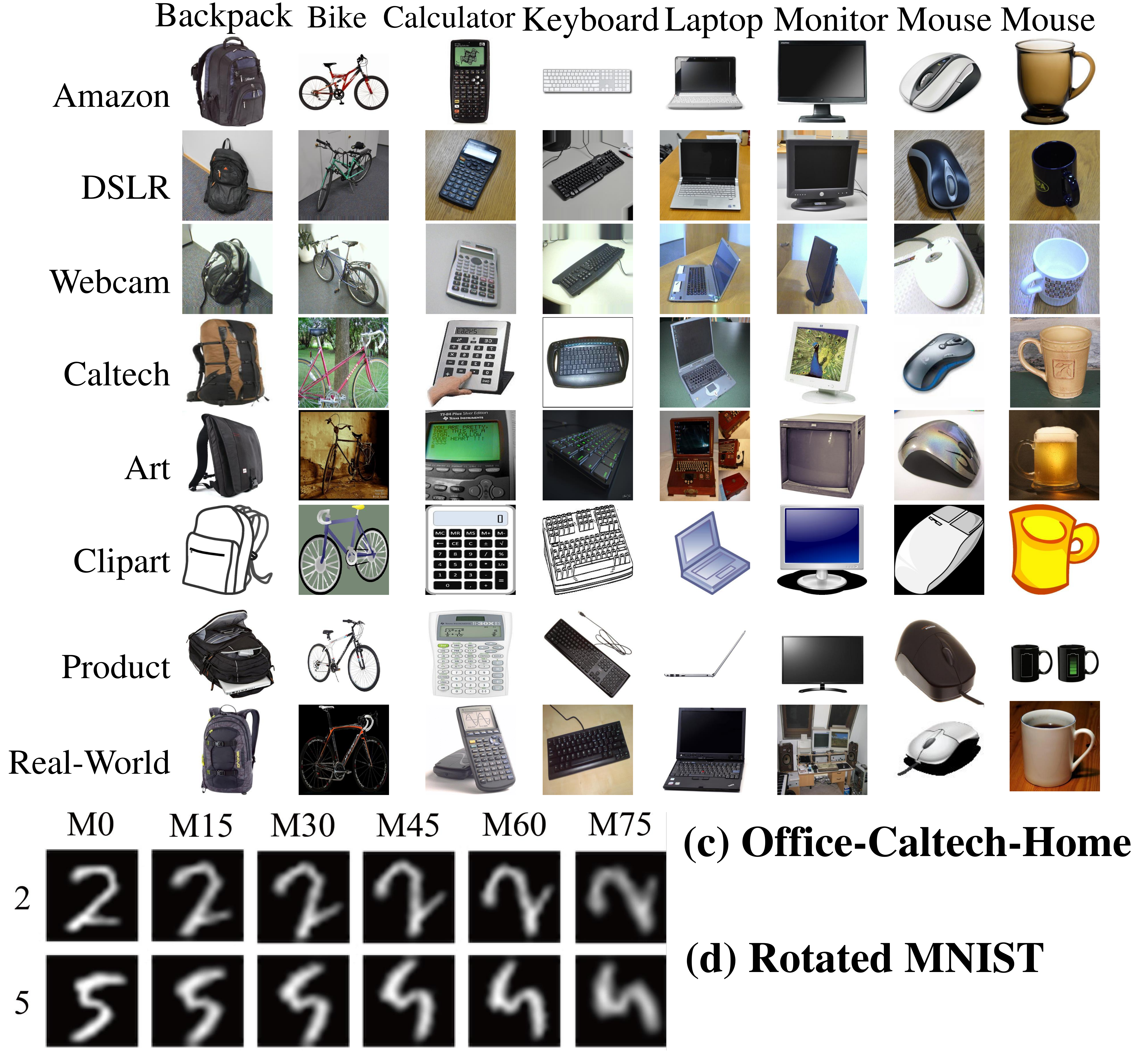}
    \caption{Some example images of the adopted datasets for experiments, i.e., PACS (a), VLCS (b), Office-Caltech-Home (c), and Rotated MNIST (d).}\label{fig-data}
\end{figure*}

\textbf{Attention mechanism.}
To encourage the cross-layer pairs with larger semantic similarity to be matched for precise semantic level alignment and domain invariance enhancement, meanwhile, weakening the pairs with less similarity for avoiding further semantic dislocation, we then introduce a semantic similarity based attention mechanism for the dynamic weight $\alpha_{l,m}$. 
Attention \cite{BahdanauCB14} is a widely adopted technique \cite{wang2017residual, zhang2019self, fu2019dual} for deciding which parts of the input features should be paid more attention to. Here, we consider the semantic inter-dependencies in both \emph{position} and \emph{channel} dimensions. Let $c$, $g$, and $w$ be the channel, height, and width of the semantic features after projection, respectively. We first reshape $M_{l}(x)'\in\mathcal{R}^{c\times{g}\times{w}}$ and $L_{m}^{h}(x)'\in\mathcal{R}^{c\times{g}\times{w}}$ to $A_{l}\in\mathcal{R}^{c\times{d}}$ and $B_{m}\in\mathcal{R}^{c\times{d}}$, respectively, where $d=g\times{w}$ is the number of pixels in an image. Then, we have a position-wise weight 
\begin{equation}
    \alpha_{l,m}^{p}=\frac{\mathrm{exp}\left(\mathrm{avg}\left(A_{l}^{\top}B_{m}\right)\right)}{\sum_{m\in\mathcal{R}}{\mathrm{exp}\left(\mathrm{avg}\left(A_{l}^{\top}B_{m}\right)\right)}},
\end{equation}
where $A_{l}^{\top}B_{m}$ is the position-wise attention map, measuring the response of $A_{l}$ to $B_{m}$, i.e., the $l$-th layer of the fused model $M_{l}$ to the $m$-th layer of the local model $L^{h}_{m}$. The operator $\mathrm{Avg}(\cdot)$ averages the attention map to a real number, and the weight $\alpha_{l,m}^{p}$ is the normalization of the average position-wise semantic similarity. 
Similarly, we then have
\begin{equation}
    \alpha_{l,m}^{c}=\frac{\mathrm{exp}\left(\mathrm{avg}\left(A_{l}B_{m}^{\top}\right)\right)}{\sum_{m\in\mathcal{R}}{\mathrm{exp}\left(\mathrm{avg}\left(A_{l}B_{m}^{\top}\right)\right)}}.
\end{equation}
$\alpha_{l,m}^{c}$ measures the channel-wise semantic similarities. We average them to get the final weight for each pair $(l,m)$:
\begin{equation}
    \alpha_{l,m}=\frac{1}{2}\left(\alpha_{l,m}^{p}+\alpha_{l,m}^{c}\right).
\end{equation}
$\alpha_{l,m}$ characterizes the inter-dependencies between the fused model $G_{l}$ and each local model $L_{m}^{h}$ in both position and channel dimensions, weighting cross-layer pairs for calibrating semantic levels and boosting model generalization. 
We adopt attention to help the model identify which feature pairs are semantically-related and which are unrelated by calculating their semantic inter-dependencies in both position and channel dimensions. Through this design, the features from different layers between the global and the local model, which have similar semantic representation, would be automatically given a larger weight for alignment, and vice versa. Therefore, we argue that our method would not mix-ups the semantic features of different layers, but align and calibrate the semantics of different layers through the dynamic attention mechanism. Extensive experiments and ablation studies also demonstrate its effectiveness. 

Since the semantic information from each domain is aggregated with others, the fused model may suffer from the \emph{catastrophic forgetting} problem \cite{mccloskey1989catastrophic, goodfellow2013empirical}, i.e., knowledge from one domain in the model is gradually forgotten when incrementally updating models with knowledge from other domains. We thus employ an auxiliary retraining loss $\mathcal{L}_{AR}^{h}$ for the model $M$ on each source dataset $\mathcal{D}^{h}$, that is,
\begin{equation}
    \mathcal{L}_{AR}^{h}=-\mathbb{E}_{(x^{h},y^{h})\in\mathcal{D}^{h}}[\sum_{c=1}^{C}\mathbbm{1}(y^{h}=c)\log{M^{c}(x^{h})}],
\end{equation}
where $M^{c}$ is the $c$-th dimension of the output of the model $M$. Then we have the calibration loss on each dataset $\mathcal{D}^{h}$: 
\begin{equation}\label{equ:cal}
    \mathcal{L}_{CB}^{h}=\lambda\mathcal{L}_{AL}^{h}+\mathcal{L}_{AR}^{h},
\end{equation}
where $\lambda$ is a hyper-parameter for semantic calibration. We minimize $\mathcal{L}_{CB}^{h}$ to optimized $M$ on each domain $h$. 
Note that optimizing the retraining loss of $L_{AR}^h$ on each client (domain) is the common practice of federated learning, like FedAvg. Therefore, we keep this loss unchanged, i.e., without using a hyper-parameter, such that we can conduct a clear ablation study and make a fair comparison with FedAvg. In addition, we try to avoid using more hyper-parameters to simplify our method. 

\subsection{Model Optimization}
We first perform local semantic acquisition by training the models $\{G^{h}\}_{h=1}^{H}$ with Equation (\ref{equ:cels}). The trained models are employed to calculate the fused model $M$ for semantic aggregation through Equation (\ref{equ:fus}). We then copy and transmit $M$ to each domain and optimize it through Equation (\ref{equ:cal}) for semantic calibration, then fuse the calibrated fused models again. 
We repeat the semantic aggregation and calibration alternately to gather semantic information from the distributed source domains and calibrate it to enhance domain-invariant information, resulting in a highly generalizable model $\hat{M}$ for inference on unknown target domains.

\textbf{Remark.} In practice, we assign the parameters of the fused model $M$ to each trained model $G^{h}$ for simultaneous semantic calibration on each domain $h$, and fuse them again.

\section{Experiments}\label{sec-exp}
\subsection{Setup}
In this section, we evaluate the proposed CSAC method for the federated domain generalization task on multiple datasets, and give in-depth ablation studies and discussions. 

\textbf{Benchmark datasets.} 
We first adopt two popular datasets of object recognition. One is \emph{PACS} \cite{li2017deeper} that covers 7 categories within 4 domains, i.e., \emph{Art}, \emph{Cartoon}, \emph{Sketch}, and \emph{Photo}. Another is \emph{VLCS} \cite{ghifary2015domain} that contains 5 classes from 4 domains, i.e., \emph{Pascal}, \emph{LabelMe}, \emph{Caltech}, and \emph{Sun}. 
A simulated digit dataset \emph{Rotated MNIST} \cite{ghifary2015domain} is then employed. It has 6 domains, i.e., \emph{M0}, \emph{M15}, \emph{M30}, \emph{M45}, \emph{M60}, and \emph{M75} through clock-wise rotation of the original images (M0) five times with fifteen degree intervals. We use 100 images per class for Rotated MNIST dataset by following \cite{ghifary2015domain, Zhao2020DomainGV}. 
We process the data by following the previous works \cite{HuangWXH20, Zhao2020DomainGV}. To evaluate the performance under the scenarios with more domains, we further construct a dataset \emph{Office-Caltech-Home} \cite{yuan2021domain} by choosing the common classes from Office-Caltech \cite{gong2012geodesic} and Office-Home \cite{venkateswara2017deep} datasets, and merge them to get 7 domains (the domain DSLR is discarded since it only contains a few images), i.e., Amazon (\emph{Am}), Webcam (\emph{We}), Caltech (\emph{Ca}), Art (\emph{Ar}), Clipart (\emph{Cl}), Product (\emph{Pr}), and Real-World (\emph{Rw}). Some representative example images of these adopted datasets are shown in Fig. \ref{fig-data}. 
We conduct leave-one-domain-out experiments, i.e., choosing one domain from each dataset to hold out as the target domain, the others are used as the (distributed) source domains. We train the model on each domain, and only transmit model parameters among domains by following \cite{mcmahan2017communication, yang2019federated, lin2020ensemble, yoonfedmix, ghosh2020efficient}.

\begin{figure*}[t]
    \centering
    \includegraphics[trim={0cm 0cm 0cm 0cm},clip,width=1.99\columnwidth]{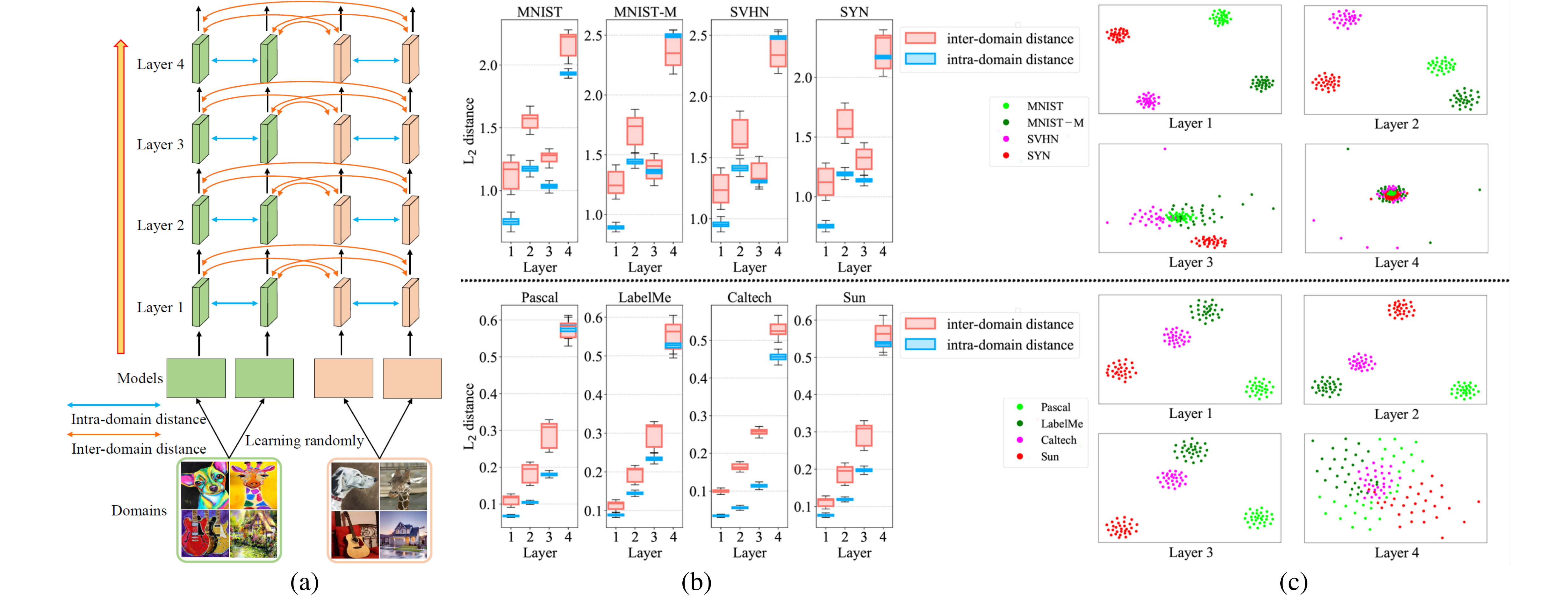}
    \caption{Results of proof-of-concept experiments. 50 models with ResNet-18 architecture are randomly run on each domain of Digits-DG (domains: MNIST, MNIST-M, SVHN, and SYN) and VLCS datasets (domains: Pascal, LabelMe, Caltech, and Sun). (a): the model parameters of the last convolution layer of the four blocks of the trained models are extracted to be layer \{1, 2, 3, 4\} and their intra- and inter-domain $L_{2}$ distance are calculated. (b): the calculated intra- and inter-domain $L_{2}$ distance of the model parameters in each layer (above: Digits-DG, below: VLCS). (c): t-SNE visualization \cite{Maaten2008VisualizingDU} of the model parameters (each point represents a trained model) in each layer (above: Digits-DG, below: VLCS).}\label{fig-toy}
\end{figure*}

\begin{table}[t]
\caption{Accuracy (\%) on PACS dataset. ``Sep.'': whether using separated source datasets. ``*'': the methods implemented by us. The best results are emphasized in bold.}
    \label{table:pacs}
    \centering
\resizebox{1\linewidth}{!}{
\begin{tabular}{l|c|cccc|c}
\toprule
Methods & Sep. & Art & Cartoon & Photo & Sketch & Average \\
\midrule
DeepAll* & \XSolidBrush 
& 78.95\scriptsize{$\pm$0.48} & 74.90\scriptsize{$\pm$1.82} & 94.08\scriptsize{$\pm$0.55} & 73.02\scriptsize{$\pm$0.80} & 80.24\scriptsize{$\pm$0.50} \\
JiGen \cite{Carlucci2019DomainGB} & \XSolidBrush & 79.42 & 75.25 & 96.03 & 71.35 & 80.51 \\
MASF \cite{dou2019domain} & \XSolidBrush & 80.29 & 77.17 & 94.99 & 71.69 & 81.04 \\
DGER \cite{Zhao2020DomainGV} & \XSolidBrush & 80.70 & 76.40 & \textbf{96.65} & 71.77 & 81.38 \\
DMG \cite{chattopadhyay2020learning} & \XSolidBrush & 76.90 & \textbf{80.38} & 92.35 & 75.21 & 81.46 \\
FACT \cite{xu2021fourier} & \XSolidBrush & \textbf{85.37} & 78.38 & 95.15 & \textbf{79.15} & \textbf{84.51} \\
Epi-FCR \cite{Li2019EpisodicTF} & \XSolidBrush & 82.1 & 77.0 & 93.9 & 73.0 & 81.5 \\
MixSyle \cite{zhou2021domain} & \XSolidBrush & 84.1 & 78.8 & 96.1 & 75.9 & 83.7 \\
EISNet \cite{Wang2020LearningFE} & \XSolidBrush & 81.89 & 76.44 & 95.93 & 74.33 & 82.15 \\ 
\midrule
FedADG \cite{zhang2021federated} & \Checkmark & 77.8\scriptsize{$\pm$0.5} & 74.7\scriptsize{$\pm$0.4} & 92.9\scriptsize{$\pm$0.3} & 79.5\scriptsize{$\pm$0.4} & 81.2 \\
FedCMI \cite{nguyen2022fedsr} & \Checkmark & 80.8\scriptsize{$\pm$0.4} & 73.7\scriptsize{$\pm$0.2} & 92.8\scriptsize{$\pm$0.5} & 79.5\scriptsize{$\pm$0.2} & 81.7 \\
FedSR \cite{nguyen2022fedsr} & \Checkmark & \textbf{83.2}\scriptsize{$\pm$0.3} & 76.0\scriptsize{$\pm$0.3} & 93.8\scriptsize{$\pm$0.5} & \textbf{81.9}\scriptsize{$\pm$0.2} & 83.7 \\
FedL2R \cite{nguyen2022fedsr} & \Checkmark & 82.2\scriptsize{$\pm$0.4} & 75.8\scriptsize{$\pm$0.3} & 92.8\scriptsize{$\pm$0.4} & 81.6\scriptsize{$\pm$0.1} & 83.1 \\
FedAvg* \cite{mcmahan2017communication} & \Checkmark & 77.49\scriptsize{$\pm$0.10} & \textbf{77.21}\scriptsize{$\pm$0.52} & 93.56\scriptsize{$\pm$0.38} & 81.19\scriptsize{$\pm$0.80} & 82.36\scriptsize{$\pm$0.44} \\
FedDG* \cite{liu2021feddg} & \Checkmark & 78.46\scriptsize{$\pm$0.20} & 75.98\scriptsize{$\pm$0.28} & 93.23\scriptsize{$\pm$0.43} & 80.92\scriptsize{$\pm$0.72} & 82.15\scriptsize{$\pm$0.35} \\
\textbf{CSAC}* (ours) & \Checkmark & 81.98\scriptsize{$\pm$0.87} & 76.41\scriptsize{$\pm$0.49} & \textbf{95.20}\scriptsize{$\pm$0.29} & 81.64\scriptsize{$\pm$0.49} & \textbf{83.81}\scriptsize{$\pm$0.33} \\
\bottomrule
\end{tabular}}
\end{table}

\begin{table}[t]
\caption{Accuracy (\%) on VLCS dataset. ``Sep.'': whether using separated source datasets. ``*'': the methods implemented by us. The best results are emphasized in bold.}
    \label{table:vlcs}
    \centering
\resizebox{1\linewidth}{!}{
\begin{tabular}{l|c|cccc|c}
\toprule
Methods & Sep. & Pascal & LabelMe & Caltech & Sun & Average \\
\midrule
\multicolumn{7}{c}{AlexNet} \\
\midrule
DeepAll* & \XSolidBrush & 71.67\scriptsize{$\pm$0.26} & 59.64\scriptsize{$\pm$0.81} & \textbf{97.48}\scriptsize{$\pm$0.14} & 67.58\scriptsize{$\pm$0.68} & 74.09\scriptsize{$\pm$0.17} \\
Epi-FCR \cite{Li2019EpisodicTF} & \XSolidBrush & 67.1 & 64.3 & 94.1 & 65.9 & 72.9 \\
JiGen \cite{Carlucci2019DomainGB} & \XSolidBrush & 70.62 & 60.90 & 96.93 & 64.30 & 73.19 \\
MASF \cite{dou2019domain} & \XSolidBrush & 69.14 & \textbf{64.90} & 94.78 & 67.64 & 74.11 \\
DGER \cite{Zhao2020DomainGV} & \XSolidBrush & \textbf{73.24} & 58.26 & 96.92 & \textbf{69.10} & 74.38 \\
EISNet \cite{Wang2020LearningFE} & \XSolidBrush & 69.83 & 63.49 & 97.33 & 68.02 & \textbf{74.67} \\
\midrule
FedAvg* \cite{mcmahan2017communication} & \Checkmark & 67.92\scriptsize{$\pm$0.26} & \textbf{60.23}\scriptsize{$\pm$0.81} & 96.85\scriptsize{$\pm$0.26} & 66.88\scriptsize{$\pm$0.22} & 72.97\scriptsize{$\pm$0.16}\\
FedDG* \cite{liu2021feddg} & \Checkmark & 67.27\scriptsize{$\pm$0.07} & 58.48\scriptsize{$\pm$0.04} & 96.83\scriptsize{$\pm$0.47} & \textbf{68.20}\scriptsize{$\pm$0.12} & 72.69\scriptsize{$\pm$0.16} \\
\textbf{CSAC}* (ours) & \Checkmark & \textbf{70.21}\scriptsize{$\pm$0.32} & 58.99\scriptsize{$\pm$0.29} & \textbf{97.13}\scriptsize{$\pm$0.35} & 67.27\scriptsize{$\pm$0.54} & \textbf{73.40}\scriptsize{$\pm$0.17} \\
\midrule
\multicolumn{7}{c}{ResNet-18} \\
\midrule
DeepAll* & \XSolidBrush & 71.40\scriptsize{$\pm$0.32} & 59.77\scriptsize{$\pm$0.95} & 97.54\scriptsize{$\pm$0.54} & 69.01\scriptsize{$\pm$0.25} & 74.43\scriptsize{$\pm$ 0.25} \\
JiGen* \cite{Carlucci2019DomainGB} & \XSolidBrush & 73.97\scriptsize{$\pm$0.21} & 61.94\scriptsize{$\pm$0.74} & 97.40\scriptsize{$\pm$1.03} & 66.90\scriptsize{$\pm$0.64} & 75.05\scriptsize{$\pm$0.26} \\
\midrule
COPA \cite{wu2021collaborative} & \Checkmark & 71.50\scriptsize{$\pm$1.05} & 61.00\scriptsize{$\pm$0.89} & 93.83\scriptsize{$\pm$0.41} & 71.72\scriptsize{$\pm$0.74} & 74.51 \\
FedAvg* \cite{mcmahan2017communication} & \Checkmark & 71.95\scriptsize{$\pm$0.06} & 63.29\scriptsize{$\pm$0.06} & 96.48\scriptsize{$\pm$0.18} & 72.37\scriptsize{$\pm$0.06} & 76.02\scriptsize{$\pm$0.08} \\
FedDG* \cite{liu2021feddg} & \Checkmark & \textbf{72.59}\scriptsize{$\pm$0.30} & 60.33\scriptsize{$\pm$0.07} & 96.70\scriptsize{$\pm$0.20} & \textbf{73.61}\scriptsize{$\pm$0.17} & 75.81\scriptsize{$\pm$0.16} \\
\textbf{CSAC}* (ours) & \Checkmark & 71.97\scriptsize{$\pm$0.56} & \textbf{63.45}\scriptsize{$\pm$0.73} & \textbf{97.24}\scriptsize{$\pm$0.57} & 72.06\scriptsize{$\pm$0.80} & \textbf{76.18}\scriptsize{$\pm$0.42} \\
\bottomrule
\end{tabular}}
\end{table}

\begin{table}[t]
\caption{Accuracy (\%) on Rotated MNIST dataset. ``Sep.'': whether using separated source datasets. ``*'': the methods implemented by us. The best results are emphasized in bold.}
    \label{table:rmnist}
    \centering
\resizebox{1\linewidth}{!}{
\renewcommand\tabcolsep{3.0pt}
\begin{tabular}{l|c|cccccc|c}
\toprule
Methods & Sep. & M0 & M15 & M30 & M45 & M60 & M75 & Average \\
\midrule
DeepAll* & \XSolidBrush 
& 86.73\scriptsize{$\pm$0.45} & 98.27\scriptsize{$\pm$0.40} & 98.63\scriptsize{$\pm$0.15} & 97.50\scriptsize{$\pm$0.89} & 97.47\scriptsize{$\pm$0.25} & 87.20\scriptsize{$\pm$0.95} & 94.30\scriptsize{$\pm$0.29} \\
CrossGrad & \XSolidBrush & 86.03 & 98.92 & 98.60 & 98.39 & 98.68 & 88.94 & 94.93 \\
MetaReg \cite{balaji2018metareg} & \XSolidBrush & 85.70 & 98.87 & 98.32 & 98.58 & 98.93 & 89.44 & 94.97 \\
FeaCri \cite{li2019feature}  & \XSolidBrush & 87.04 & \textbf{99.53} & \textbf{99.41} & \textbf{99.52} & 99.23 & \textbf{91.52} & 96.04 \\
DGER \cite{Zhao2020DomainGV}  & \XSolidBrush & \textbf{90.09} & 99.24 & 99.27 & 99.31 & \textbf{99.45} & 90.81 & \textbf{96.36} \\
\midrule
FedAvg* \cite{mcmahan2017communication}  & \Checkmark & 82.60\scriptsize{$\pm$0.44} & 98.56\scriptsize{$\pm$0.27} & \textbf{98.97}\scriptsize{$\pm$0.29} & 93.66\scriptsize{$\pm$0.03} & 95.78\scriptsize{$\pm$0.27} & 86.30\scriptsize{$\pm$0.10} & 92.65\scriptsize{$\pm$0.08} \\
FedDG* \cite{liu2021feddg}  & \Checkmark & 73.07\scriptsize{$\pm$0.67} & 94.37\scriptsize{$\pm$1.03} & 95.60\scriptsize{$\pm$0.19} & 89.43\scriptsize{$\pm$0.38} & 94.61\scriptsize{$\pm$0.39} & 84.50\scriptsize{$\pm$0.10} & 88.60\scriptsize{$\pm$0.30} \\
\textbf{CSAC}* (ours) & \Checkmark & \textbf{84.57}\scriptsize{$\pm$0.31} & \textbf{98.87}\scriptsize{$\pm$0.23} & 98.63\scriptsize{$\pm$0.15} & \textbf{95.06}\scriptsize{$\pm$0.48} & \textbf{96.57}\scriptsize{$\pm$0.40} & \textbf{90.73}\scriptsize{$\pm$0.25} & \textbf{94.07}\scriptsize{$\pm$0.02} \\
\bottomrule
\end{tabular}}
\end{table}

\begin{figure}[t]
    \centering
    \includegraphics[trim={2cm 0cm 2cm 1cm},clip,width=0.9\columnwidth]{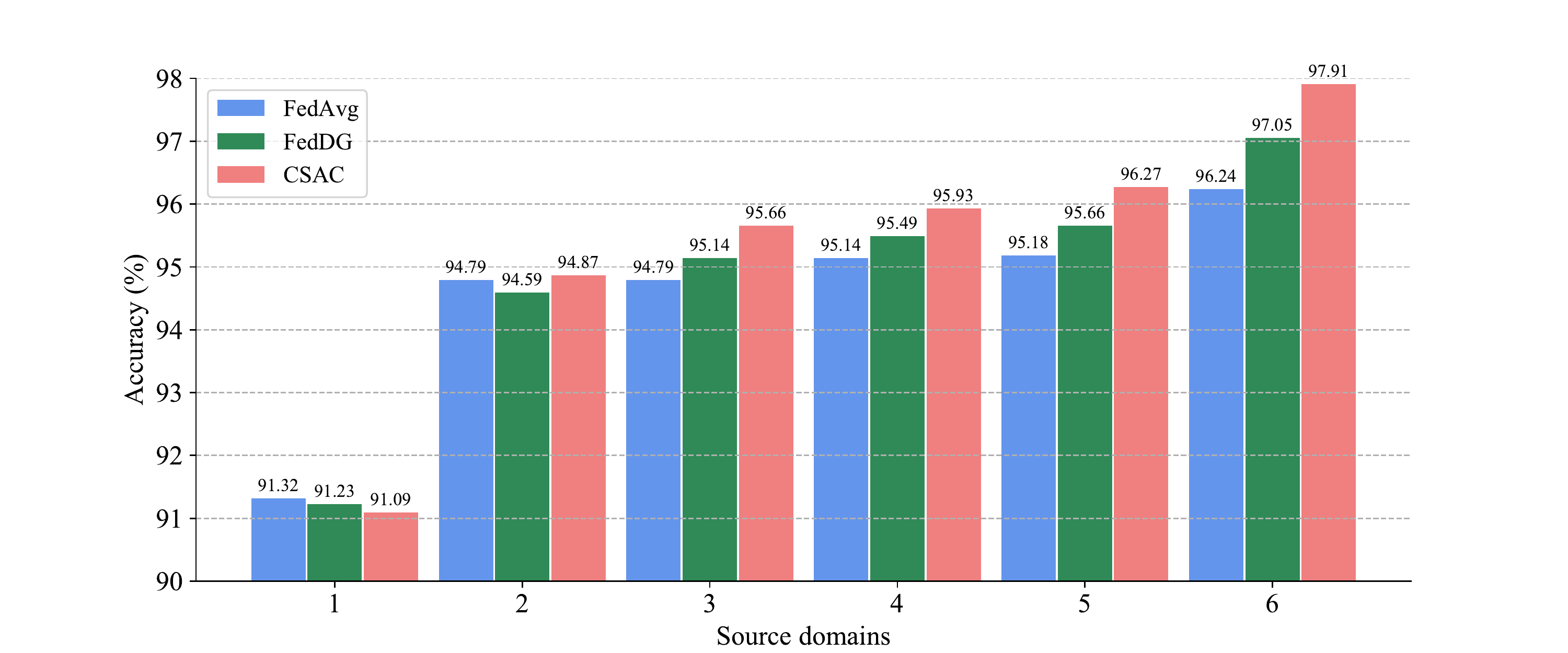}
    \caption{Results on Office-Caltech-Home dataset with 7 domains. We let domain Rw be the target domain, and add one source domain each time from a source domain set \{Am, Cl, Pr, We, Ar, Ca\}, i.e., using 1 source domain: \{Am\}, using 2 source domains: \{Am, Cl\}, and so on.}\label{fig:office-caltech-home}
\end{figure}

\begin{table*}[t]
\caption{Effect of semantic aggregation with semantic divergence (strategy) and $L_{2}$ distance (metric). The best results are emphasized in bold.}
    \label{table:abl-agg}
    \centering
\resizebox{1\linewidth}{!}{
\begin{tabular}{l|l|cccc|c|cccc|c}
\toprule
\multirow{2}{*}{Type} & \multirow{2}{*}{Case} & \multicolumn{5}{c}{PACS} & \multicolumn{5}{|c}{VLCS} \\
\cline{3-12}
& & Art & Cartoon & Photo & Sketch & Average & Pascal & LabelMe & Caltech & Sun & Average \\
\midrule
\multirow{2}{*}{Strategy} & Semantic similarity & 79.22\scriptsize{$\pm$0.19} & 72.44\scriptsize{$\pm$0.53} & 93.00\scriptsize{$\pm$0.14} & 74.34\scriptsize{$\pm$0.37} & 79.75\scriptsize{$\pm$0.22} & 70.80\scriptsize{$\pm$0.16} & 62.86\scriptsize{$\pm$0.39} & 96.61\scriptsize{$\pm$0.26} & 69.49\scriptsize{$\pm$1.03} & 74.94\scriptsize{$\pm$0.46} \\
& Semantic average & 80.25\scriptsize{$\pm$0.36} & 74.54\scriptsize{$\pm$0.44} & 93.74\scriptsize{$\pm$0.15} & 77.88\scriptsize{$\pm$0.69} & 81.85\scriptsize{$\pm$0.07} & \textbf{72.44}\scriptsize{$\pm$0.17} & 61.20\scriptsize{$\pm$0.32} & 96.54\scriptsize{$\pm$0.30} & 70.83\scriptsize{$\pm$0.45} & 75.25\scriptsize{$\pm$0.21} \\
\midrule
\multirow{2}{*}{Metric} & Cosine distance & 80.89\scriptsize{$\pm$0.23} & 73.57\scriptsize{$\pm$0.41} & 94.43\scriptsize{$\pm$0.03} & 76.59\scriptsize{$\pm$1.31} & 81.37\scriptsize{$\pm$0.41} & 72.27\scriptsize{$\pm$0.11} & 62.94\scriptsize{$\pm$0.44} & 97.05\scriptsize{$\pm$0.15} & 71.41\scriptsize{$\pm$0.06} & 75.92\scriptsize{$\pm$0.11} \\
% \cline{2-12}
& $L_{1}$ distance & 81.75\scriptsize{$\pm$0.51} & 74.93\scriptsize{$\pm$0.09} & 94.45\scriptsize{$\pm$0.28} & 77.63\scriptsize{$\pm$0.42} & 82.19\scriptsize{$\pm$0.27} & 72.16\scriptsize{$\pm$0.16} & \textbf{64.11}\scriptsize{$\pm$0.28} & 96.64\scriptsize{$\pm$0.13} & 71.67\scriptsize{$\pm$0.21} & 76.15\scriptsize{$\pm$0.04} \\
\midrule
\multicolumn{2}{c|}{\textbf{CSAC}} & \textbf{81.98}\scriptsize{$\pm$0.87} & \textbf{76.41}\scriptsize{$\pm$0.49} & \textbf{95.20}\scriptsize{$\pm$0.29} & \textbf{81.64}\scriptsize{$\pm$0.49} & \textbf{83.81}\scriptsize{$\pm$0.33} & 71.97\scriptsize{$\pm$0.56} & 63.45\scriptsize{$\pm$0.73} & \textbf{97.24}\scriptsize{$\pm$0.57} & \textbf{72.06}\scriptsize{$\pm$0.80} & \textbf{76.18}\scriptsize{$\pm$0.42} \\
\bottomrule
\end{tabular}}
\end{table*}

\begin{table*}[t]
\caption{Effect of semantic calibration with cross-layer alignment (strategy) and MMD discrepancy (metric). The best results are emphasized in bold.}
    \label{table:abl-cal}
    \centering
\resizebox{1\linewidth}{!}{
\begin{tabular}{l|l|cccc|c|cccc|c}
\toprule
\multirow{2}{*}{Type} & \multirow{2}{*}{Case} & \multicolumn{5}{c}{PACS} & \multicolumn{5}{|c}{VLCS} \\
\cline{3-12}
& & Art & Cartoon & Photo & Sketch & Average & Pascal & LabelMe & Caltech & Sun & Average \\
\midrule
\multirow{6}{*}{Strategy} & Without alignment & 80.34\scriptsize{$\pm$0.24} & 76.07\scriptsize{$\pm$0.14} & \textbf{95.45}\scriptsize{$\pm$0.49} & 81.11\scriptsize{$\pm$0.12} & 83.24\scriptsize{$\pm$0.15} & 70.11\scriptsize{$\pm$1.21} & \textbf{65.73}\scriptsize{$\pm$2.31} & 97.03\scriptsize{$\pm$0.15} & 71.39\scriptsize{$\pm$0.16} & 76.06\scriptsize{$\pm$0.75} \\
& Same-layer alignment & 80.49\scriptsize{$\pm$0.23} & 74.19\scriptsize{$\pm$0.11} & 94.85\scriptsize{$\pm$0.55} & 80.34\scriptsize{$\pm$0.44} & 82.46\scriptsize{$\pm$0.17} & \textbf{73.51}\scriptsize{$\pm$0.21} & 62.26\scriptsize{$\pm$0.14} & 97.38\scriptsize{$\pm$0.08} & 71.38\scriptsize{$\pm$0.14} & 76.14\scriptsize{$\pm$0.12} \\
\cline{2-12}
& Without attention (position) & 79.57\scriptsize{$\pm$0.44} & 73.57\scriptsize{$\pm$0.61} & 94.46\scriptsize{$\pm$0.12} & 79.66\scriptsize{$\pm$0.47} & 81.82\scriptsize{$\pm$}0.16 & 72.57\scriptsize{$\pm$0.12} & 63.85\scriptsize{$\pm$0.78} & 96.09\scriptsize{$\pm$0.26} & 69.16\scriptsize{$\pm$0.21} & 75.42\scriptsize{$\pm$0.31} \\
& Without attention (channel) & 79.80\scriptsize{$\pm$0.58} & 73.77\scriptsize{$\pm$0.03} & 94.43\scriptsize{$\pm$0.09} & 79.32\scriptsize{$\pm$0.19} & 81.83\scriptsize{$\pm$0.14} & 72.40\scriptsize{$\pm$0.52} & 63.54\scriptsize{$\pm$0.52} & 95.92\scriptsize{$\pm$0.30} & 69.24\scriptsize{$\pm$0.12} & 75.27\scriptsize{$\pm$0.20} \\
& Without attention & 80.35\scriptsize{$\pm$0.87} & 76.27\scriptsize{$\pm$0.11} & 93.59\scriptsize{$\pm$0.24} & 77.88\scriptsize{$\pm$1.53} & 82.02\scriptsize{$\pm$0.58} & 71.63\scriptsize{$\pm$0.06} & 64.48\scriptsize{$\pm$2.46} & 96.27\scriptsize{$\pm$0.40} & 67.67\scriptsize{$\pm$1.03} & 75.01\scriptsize{$\pm$0.44} \\
\cline{2-12}
& Without label smoothing  & 81.78\scriptsize{$\pm$1.00} & 76.37\scriptsize{$\pm$0.43} & 95.00\scriptsize{$\pm$0.71} & 81.46\scriptsize{$\pm$0.80} & 83.65\scriptsize{$\pm$0.52} & 71.70\scriptsize{$\pm$0.78} & 63.33\scriptsize{$\pm$0.82} & 97.17\scriptsize{$\pm$0.61} & 71.34\scriptsize{$\pm$1.18} & 75.89\scriptsize{$\pm$0.54} \\
& Without cross-entropy & 81.35\scriptsize{$\pm$0.09} & 76.04\scriptsize{$\pm$0.17} & 94.98\scriptsize{$\pm$0.39} & \textbf{82.60}\scriptsize{$\pm$0.50} & 83.74\scriptsize{$\pm$0.13} & 72.12\scriptsize{$\pm$0.23} & 62.70\scriptsize{$\pm$0.16} & \textbf{97.55}\scriptsize{$\pm$0.21} & 71.76\scriptsize{$\pm$0.23} & 76.03\scriptsize{$\pm$0.10} \\
\midrule
Metric & Mean Square Error (MSE) & 77.08\scriptsize{$\pm$0.28} & 71.05\scriptsize{$\pm$1.26} & 94.61\scriptsize{$\pm$0.14} & 75.99\scriptsize{$\pm$0.33} & 79.68\scriptsize{$\pm$0.49} & 72.10\scriptsize{$\pm$0.79} & 62.17\scriptsize{$\pm$0.19} & 96.47\scriptsize{$\pm$0.35} & 71.09\scriptsize{$\pm$0.14} & 75.46\scriptsize{$\pm$0.26} \\
\midrule
\multicolumn{2}{c|}{\textbf{CSAC}} & \textbf{81.98}\scriptsize{$\pm$0.87} & \textbf{76.41}\scriptsize{$\pm$0.49} & 95.20\scriptsize{$\pm$0.29} & 81.64\scriptsize{$\pm$0.49} & \textbf{83.81}\scriptsize{$\pm$0.33} & 71.97\scriptsize{$\pm$0.56} & 63.45\scriptsize{$\pm$0.73} & 97.24\scriptsize{$\pm$0.57} & \textbf{72.06}\scriptsize{$\pm$0.80} & \textbf{76.18}\scriptsize{$\pm$0.42} \\
\bottomrule
\end{tabular}}
\end{table*}

\begin{figure}[t]
    \centering
    \includegraphics[trim={0cm 0cm 0cm 0cm},clip,width=0.45\columnwidth]{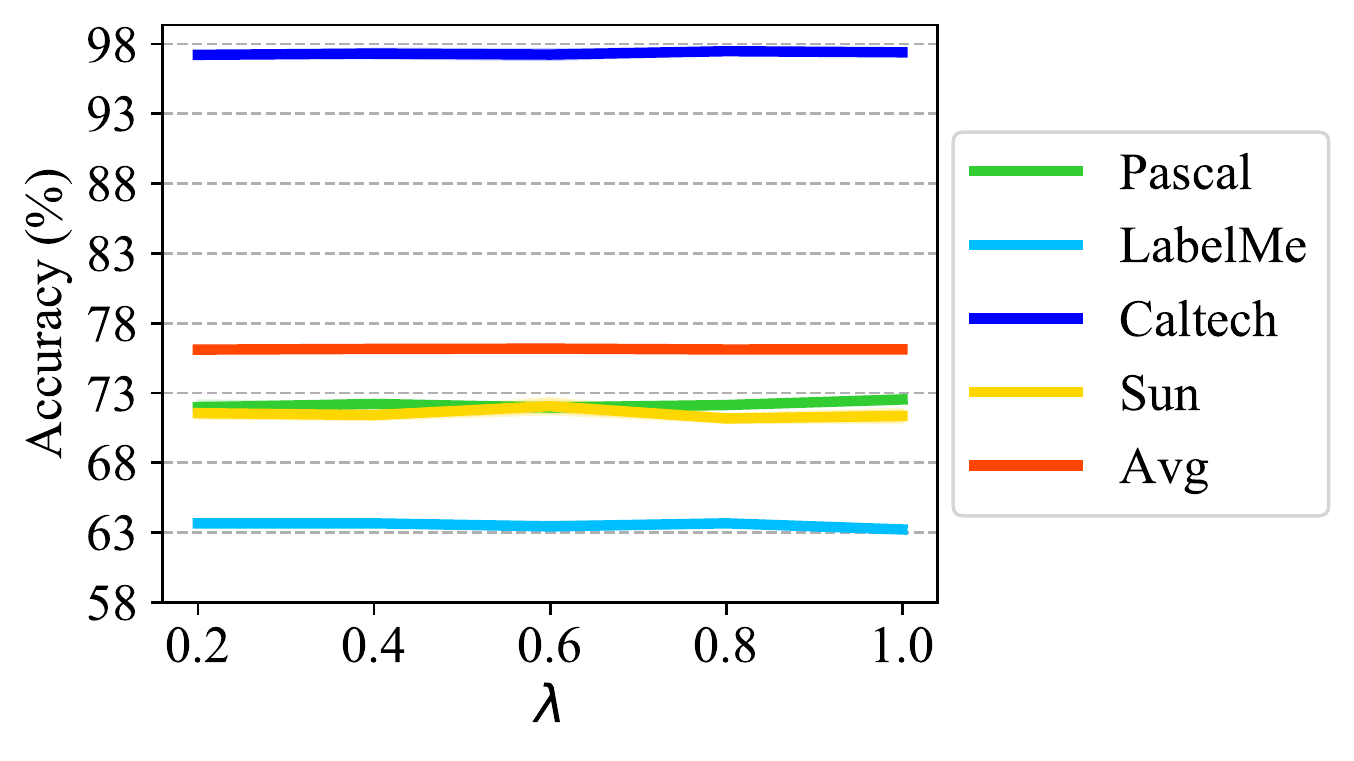}
    \includegraphics[trim={0cm 0cm 0cm 0cm},clip,width=0.45\columnwidth]{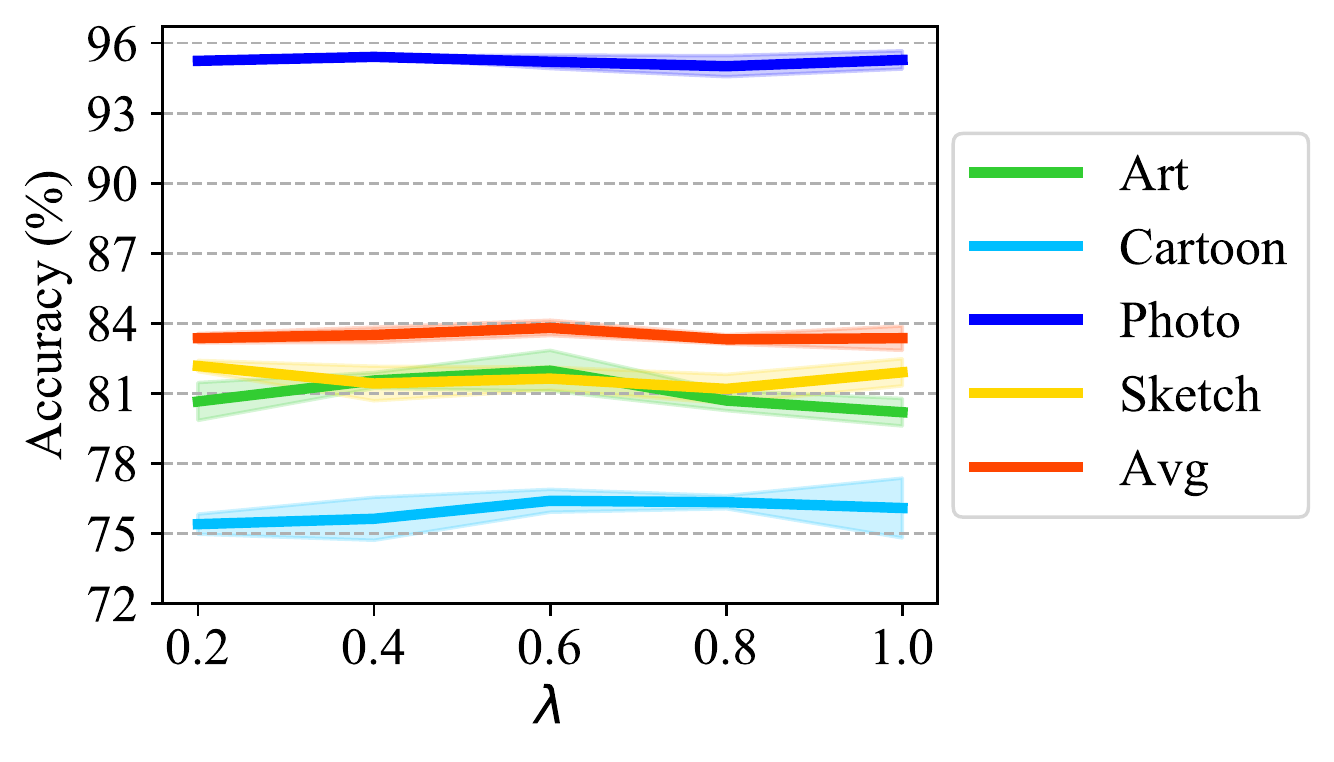}
    \caption{Sensitivity analysis of the hyper-parameter $\lambda$ for semantic calibration on VLCS (left) and PACS (right) datasets.}\label{fig:sen}
\end{figure}

\begin{table*}[t]
\caption{Run time (hours) on PACS and VLCS datasets.}\label{table:runtime}
    \centering
\resizebox{0.8\linewidth}{!}{
\begin{tabular}{c|cccc|c|cccc|c}
\toprule
\multirow{2}{*}{Methods}& \multicolumn{5}{c|}{PACS} & \multicolumn{5}{c}{VLCS} \\
\cline{2-11}
 & Photo & Art & Cartoon & Sketch & Average & Pascal & LabelMe & Caltech & Sun & Average \\
\midrule
FedAvg \cite{mcmahan2017communication} & 3.85 & 3.95 & 4.03 & 4.32 & 4.04 & 3.03 & 3.10 & 3.08 & 3.11 & 3.08 \\
FedDG \cite{liu2021feddg} & 11.65 & 11.84 & 11.92 & 12.01 & 11.86 & 10.81 & 10.43 & 10.50 & 10.19 & 10.48 \\
\textbf{CSAC} (ours) & 4.81 & 4.81 & 4.48 & 4.34 & 4.61 & 3.61 & 3.61 & 3.57 & 3.59 & 3.60 \\
\bottomrule
\end{tabular}}
\end{table*}

\begin{figure*}[t]
    \centering
    \includegraphics[trim={0cm 0cm 0cm 0cm},clip,width=1.8\columnwidth]{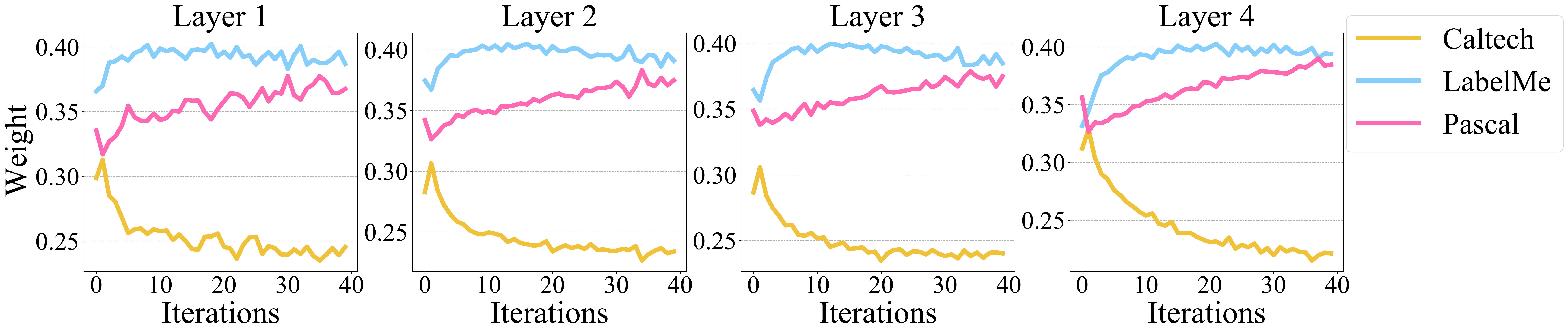}
    \includegraphics[trim={0cm 0cm 0cm 0cm},clip,width=1.8\columnwidth]{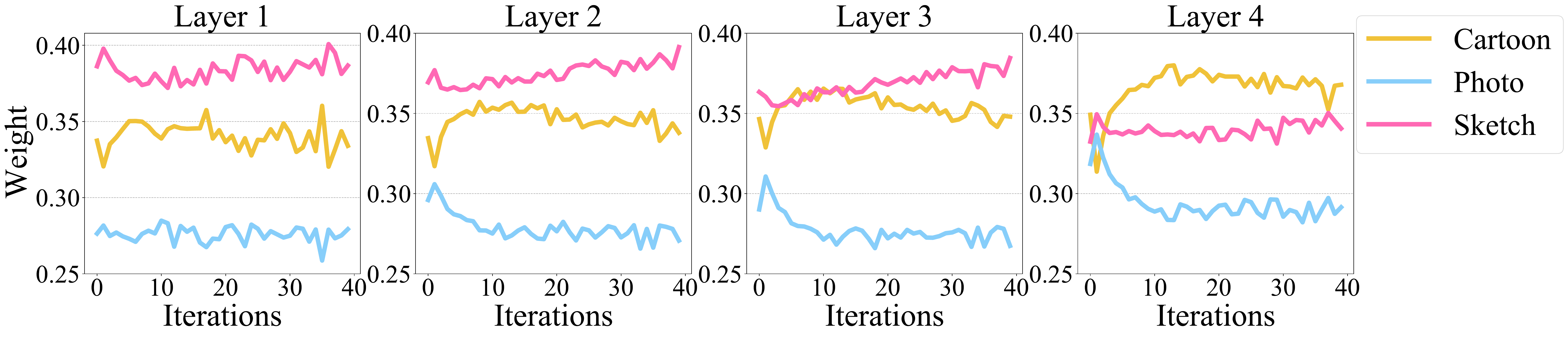}
    \caption{Semantic divergence based weight for the last layer of the four blocks of ResNet-18 of each trained model (marked with the corresponding domain) during training on VLCS (above) and PACS (below).}\label{fig:weight}
\end{figure*}

\begin{figure*}[t]
    \centering
    \includegraphics[trim={0cm 0cm 0cm 0cm},clip,width=1.99\columnwidth]{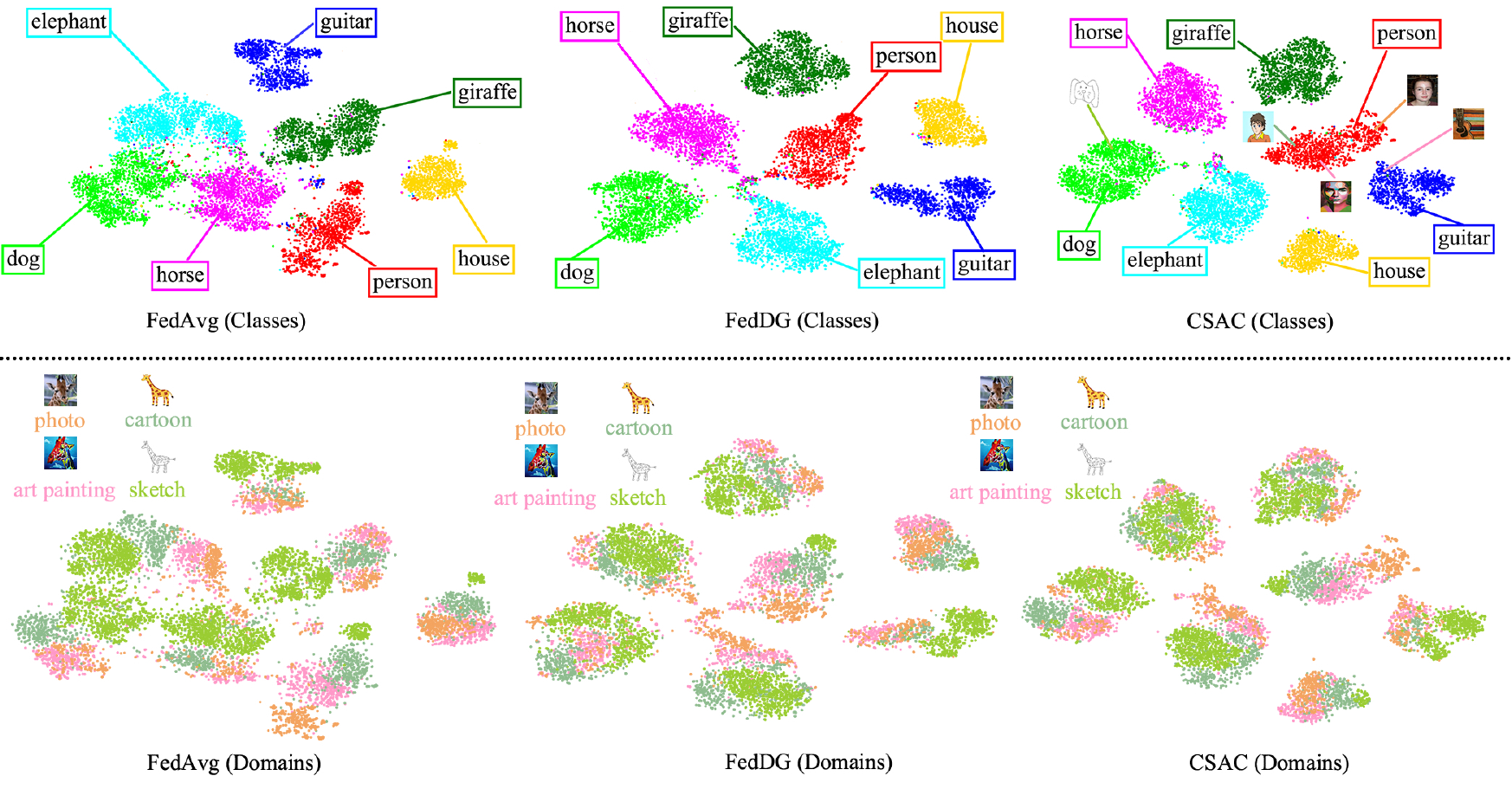}
    \caption{T-SNE visualization of the learned semantic feature distributions of the data points on PACS dataset (target domain is photo). Different colors represent different classes (above) or domains (below). }\label{fig:tsne}
\end{figure*}

\begin{figure*}[t]
    \centering
    \includegraphics[trim={0cm 0cm 0cm 0cm},clip,width=0.48\columnwidth]{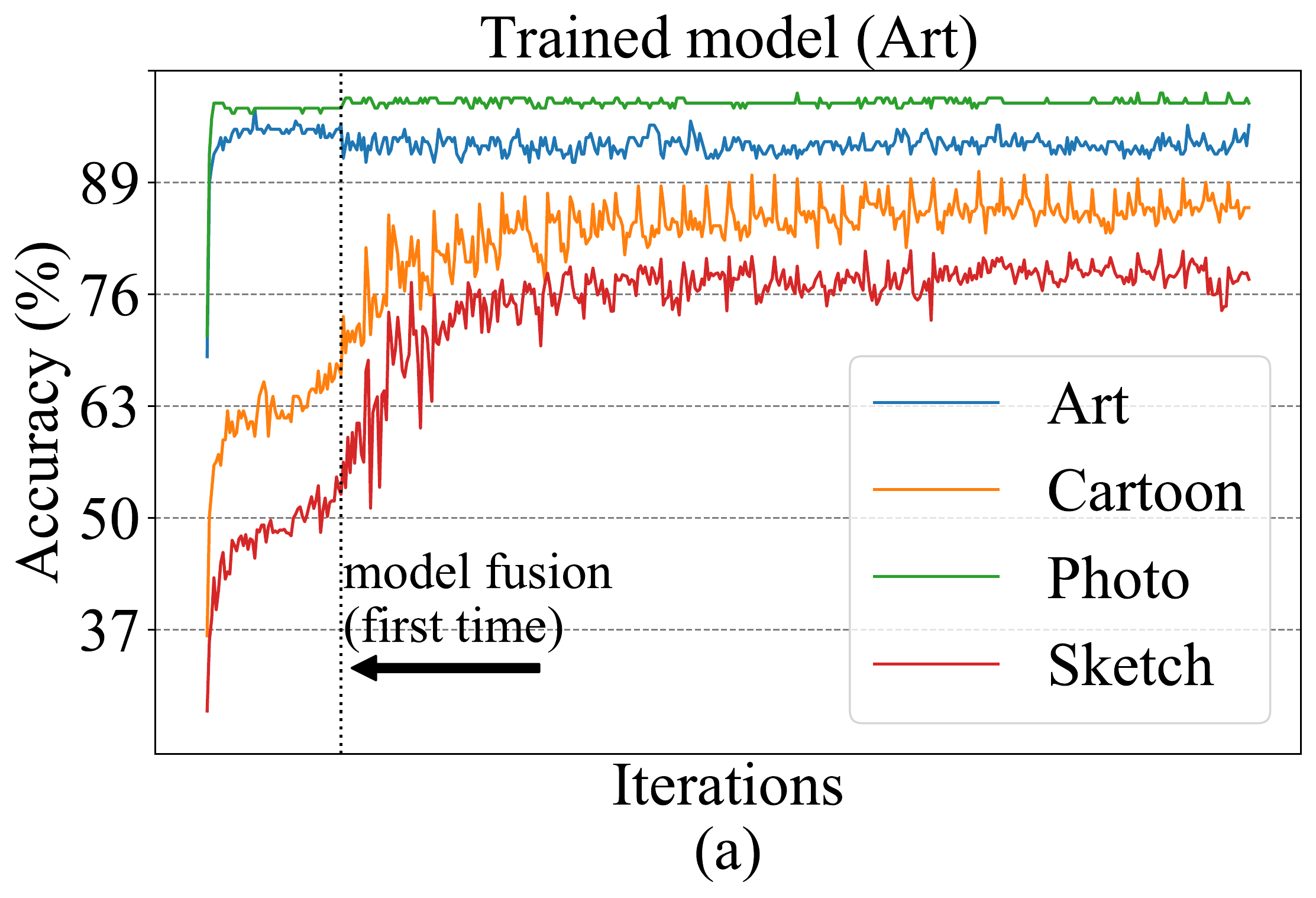}
    \includegraphics[trim={0cm 0cm 0cm 0cm},clip,width=0.48\columnwidth]{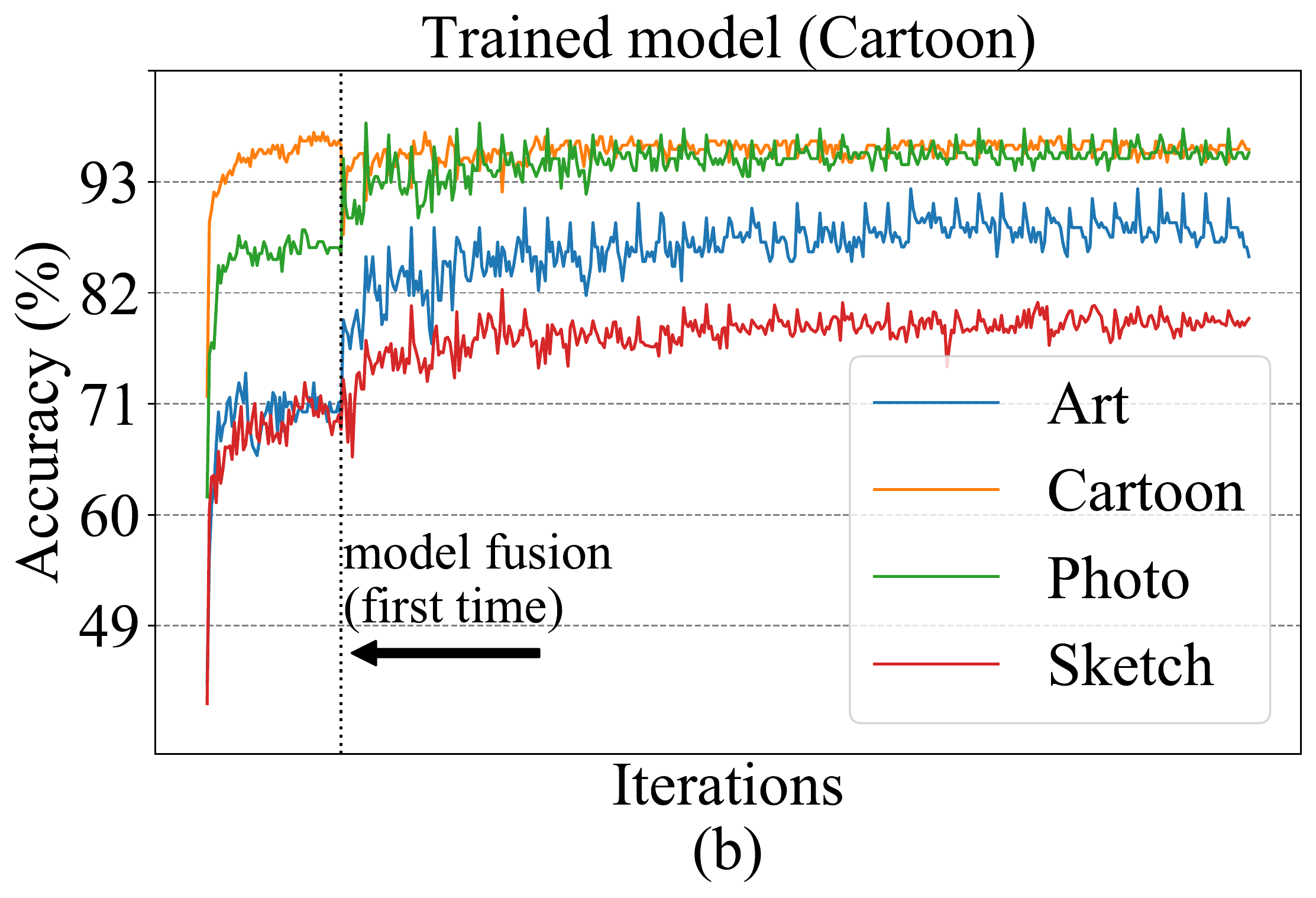}
    \includegraphics[trim={0cm 0cm 0cm 0cm},clip,width=0.48\columnwidth]{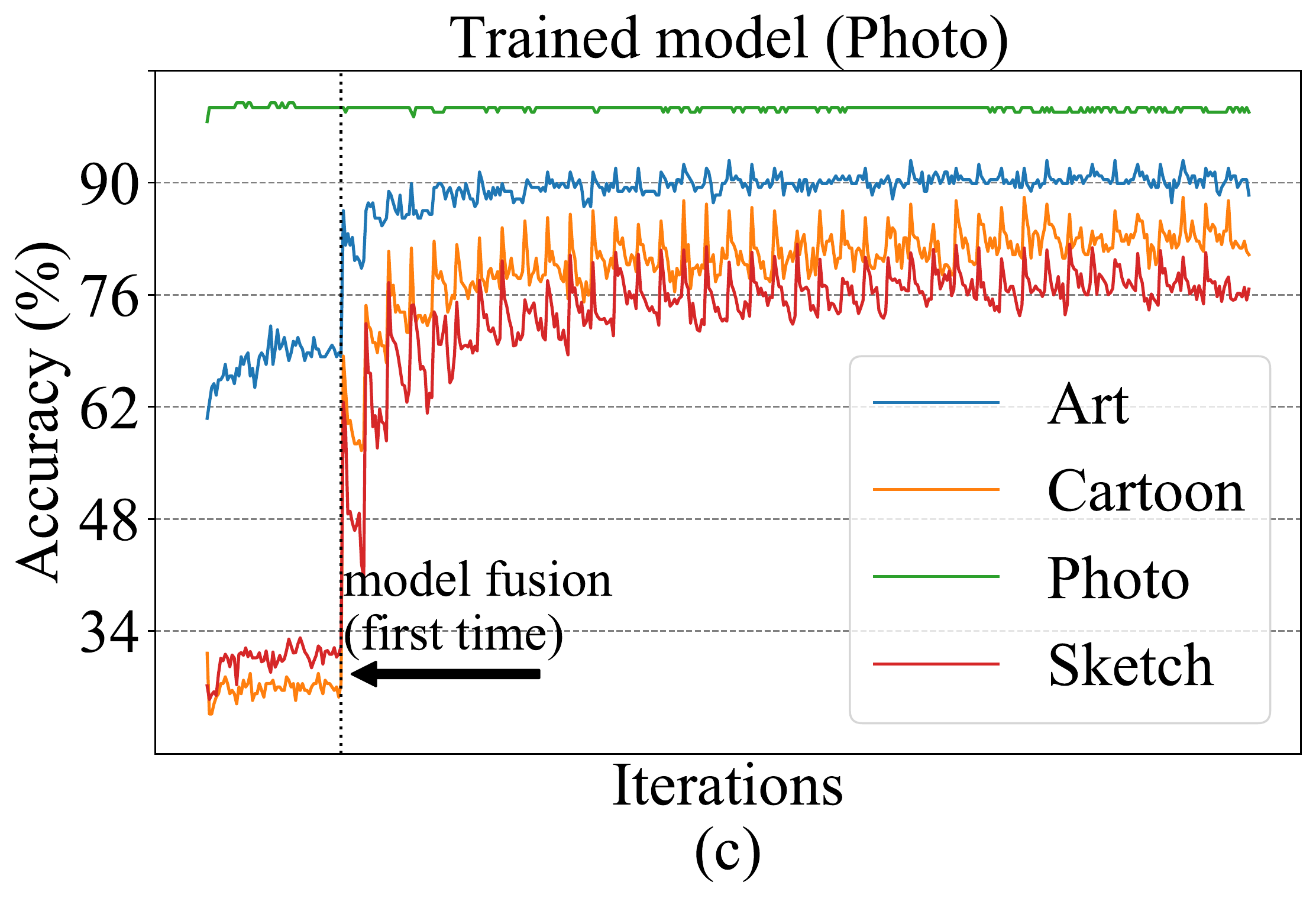}
    \includegraphics[trim={0cm 0cm 0cm 0cm},clip,width=0.48\columnwidth]{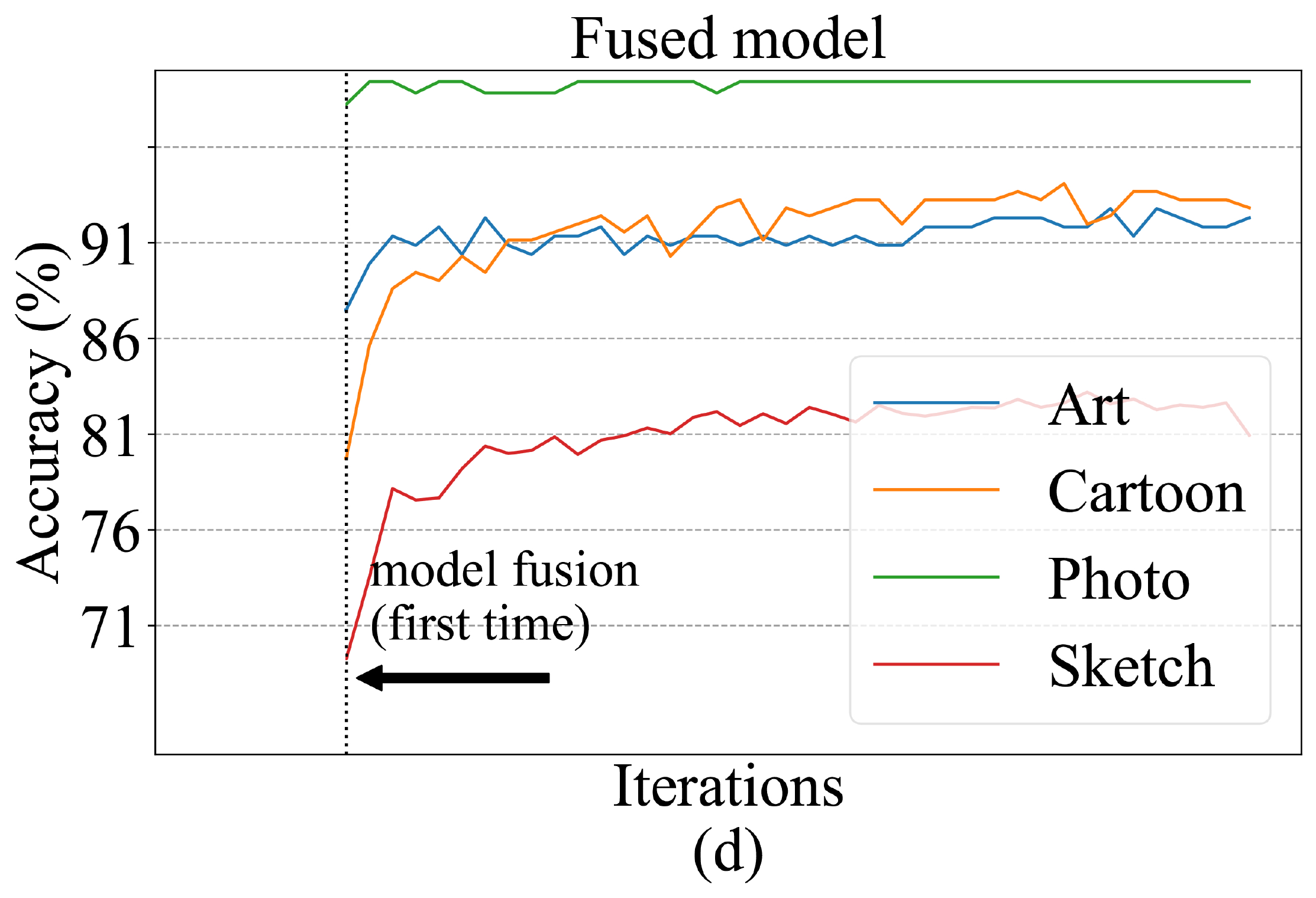}
    \caption{Accuracy (on all the domains) of the model trained on domain Art (a), Cartoon (b), and Photo (c), and the fused model (d), during the training process, i.e., semantic acquisition and the repeat of semantic aggregation and calibration. The adopted dataset (target domain): PACS (Sketch).}\label{fig:acc-train}
\end{figure*}

\textbf{Baseline methods.} We compare our method CSAC against the representative federated learning method \emph{FedAvg} \cite{mcmahan2017communication} and the federated learning based generalizable model learning method \emph{FedDG} \cite{liu2021feddg} in the separated domain generalization task. We also show the performance of the state-of-the-art DG methods (see Table \ref{table:pacs}, \ref{table:vlcs}, and \ref{table:rmnist}) introduced in the Sec. \ref{sec-rel} for the domain generalization task with shared source data. 
Following the previous works \cite{dou2019domain, Zhao2020DomainGV, Carlucci2019DomainGB}, we implement a baseline method \emph{DeepAll} by employing the fusion of the shared source datasets for model training. 

\textbf{Implementation details.} We use the pretrained ResNet-18 network \cite{he2016deep} for PACS, VLCS, and Office-Caltech-Home datasets and also use the AlexNet network \cite{KrizhevskySH17} for VLCS by following \cite{Carlucci2019DomainGB, dou2019domain, HuangWXH20}. We use standard MNIST CNN architecture with two convolution layers and two fully-connected layers for Rotated MNIST dataset by following \cite{ghifary2015domain, Zhao2020DomainGV}. We extract the convolution layers of the last three blocks of ResNet-18, or the last three convolution layers of AlexNet, or the last two convolution layers of MNIST CNN, as the layer set for semantic calibration. 
We implement the methods according to their public code, where the boundary component (it is used for segmentation task) of FedDG is discarded for fair comparison. 
We use SGD optimizer with learning rate 0.01 and momentum 0.5 for ResNet-18 and MNIST CNN, and learning rate 0.001 for AlexNet. 
The training epochs for semantic acquisition are set to 30, collaboration rounds for aggregation and calibration are set to 40, for all the datasets. In each round, the calibration epochs are set to 5 and 10 for the Rotated MNIST and the other datasets, respectively. 
The hyper-parameter $\lambda$ is set to 0.6 for all the experiments, its sensitivity is further analyzed. 
We run the experiments on a device with CPU Xeon Gold 6254 $\times$ 2, and GPU Nvidia RTX 2080 TI $\times$ 4.
We report the mean and standard error of the classification accuracy over five runs with random seeds for the experiments implemented by us (marked with *). And we cite other results of the DG methods from the published papers like the previous works.

\subsection{Proof-of-Concept Experiments}
We first provide proof-of-concept experiments to verify our hypothesis: the deep models extract semantic information layer-by-layer, and the model parameters in each layer are related to the corresponding level as well as the training data distribution. We randomly run 50 models with ResNet-18 architecture on each domain of VLCS \cite{ghifary2015domain} and Digits-DG \cite{zhou2020deep} datasets. After training, we extract the model parameters of the last convolution layer of the four blocks of the trained models to be layer \{1, 2, 3, 4\}. We calculate \emph{intra-domain distance}, i.e., the  $L_{2}$ distance of the model parameters between each pair of the models trained on the same domain, and \emph{inter-domain distance}, i.e., the $L_{2}$ distance of the model parameters between each pair of the models trained on different domains, as shown in Fig. \ref{fig-toy} (a). From the distance results in Fig. \ref{fig-toy} (b), we observe that the models trained on the same domain have closer parameter distance than the models trained on different domains, which is also verified by the t-SNE visualization \cite{Maaten2008VisualizingDU} in Fig. \ref{fig-toy} (c) that the points of the model parameters from the same domain gather together. It verifies that the model parameters are related to the distributions of training data. 
Meanwhile, we find that the inter-domain distance becomes closer to the intra-domain distance in the high layers in Fig. \ref{fig-toy} (b) and the points with different colors cluster together in the high layers in Fig. \ref{fig-toy} (c). It indicates that the model parameters are not only related to the training data distributions but also the corresponding semantic level, since the lower semantic level is more related to the data distribution while the higher level is more related to the object categories that is invariant to the domains. We argue that it is also the latent assumption of domain generalization task that one can extract high-level discriminative yet domain-agnostic semantics for training a highly generalizable model.

\subsection{Main Results}
We first report the results on PACS dataset in Table \ref{table:pacs}. We observe that our method CSAC achieves the highest average accuracy for the federated domain generalization task. Moreover, CSAC with separated source data even outperforms the domain generalization methods (except FACT) with shared source data on the average accuracy. It shows the effectiveness of our collaborative semantic aggregation and calibration strategy. Moreover, it indicates that we may learn a generalizable model by sharing information among domains through the model parameters. In this way, we can improve generalization performance of the model under careful data privacy protection, which is important for many privacy-sensitive real-world scenarios.

We further use models with AlextNet and ResNet-18 architecture for the experiments on VLCS dataset and report the results in Table \ref{table:vlcs}. In the experiments of the two network architecture, our method CSAC surpasses the FedAvg and FedDG methods. It shows excellent generalization learning ability of CSAC with separated source data. With the AlexNet network architecture, CSAC slightly outperforms Epi-FCR and JiGen methods with shared source data. However, by using the larger network ResNet-18, CSAC performs better than DeepAll and JiGen. We attribute it to the semantic calibration process and the attention mechanism which may need large network to show their advantages.  

We report the results on Rotated MNIST dataset in Table \ref{table:rmnist}. It demonstrates that our method CSAC defeats FedAvg and FedDG methods for the distributed domain generalization task. We also observe that CSAC achieves slightly worse performance than the domain generalization methods with shared data. It is probably because we use a much smaller network here for the Rotated MNIST dataset. It is similar to our previous conclusion that CSAC needs large network to unleash the potential of domain invariance learning. 

In order to evaluate the methods under the scenarios with more domains, we further conduct experiments on the Office-Caltech-Home dataset with 7 domains. 
We let domain Rw be the target domain, and add one source domain each time from a source domain set \{Am, Cl, Pr, We, Ar, Ca\}, i.e., using 1 source domain: \{Am\}, using 2 source domains: \{Am, Cl\}, and so on. The results are shown in Fig. \ref{fig:office-caltech-home}. We have two observations: (1) Our method CSAC surpasses other methods when given more than two source domains. (2) Giving more source domains enable CSAC to achieve much better performance. We argue that it is because the adequate semantic information provided by multiple source domains facilitate the semantic level alignment and invariance enhancement in the CSAC framework.

\subsection{In-Depth Ablation Studies}
\textbf{Semantic aggregation.}
Table \ref{table:abl-agg} reports ablation results for semantic aggregation. By replacing the strategy with semantic similarity (larger weights for the models that are closer to the average distribution) and average (equal weights), we find that it is important to pay more attention to the domain with the semantic distribution far from the others for fairly absorbing knowledge from all the source domain, facilitating valid domain invariance learning. The experiments with other metrics show the effectiveness of the used $L_{2}$ distance. 

\textbf{Semantic calibration.}
Table \ref{table:abl-cal} reports the ablation results for semantic calibration. 
We compare the results without alignment and using the same-layer alignment and find that it is necessary to consider cross-layer semantic relationships for addressing the semantic dislocation problem. 
By conducting the attention ablations, we demonstrate the attention mechanism with both position and channel inter-dependency consideration is important for precise semantic level alignment. 
The label smoothing is showed useful for a final generalizable model learning, which is may because it leads to more smooth models for stable model fusion. 
Cross-entropy displays its importance for catastrophic forgetting. 
The MMD metric is more effective for alignment than the MSE, which may also be the reason that MMD is widely adopted in alignment-based domain adaptation works. 

\textbf{Sensitivity analysis.}
Fig. \ref{fig:sen} shows that CSAC is generally robust to the weight of semantic calibration, i.e., hyper-parameter $\lambda$. CSAC might be practical and effective without the time-consuming hyper-parameter fine-tuning. 

\subsection{Run Time}
We report the run time of the methods (implemented locally) in Table \ref{table:runtime}. FedDG is computationally inefficient by using about three times the run time of FedAvg and CSAC, which is may because of the time-consuming process of the distribution bank building. Besides, FedDG transmits the bank to all the domains, which needs high communication costs and might increase the risks of privacy leakage (although we can not verify it with experiments).

\subsection{Why Does CSAC Work?}
In Fig. \ref{fig:weight}, we observe that the weight curves have the similar trend, i.e., the four extracted layers of each model have the similar semantic divergence to others, which indirectly verifies our hypothesis that parameters are related to the data. The model with divergent semantics is given large weight layer-by-layer for adequate semantic gathering, facilitating the semantic calibration as shown in ablation studies. 

Fig. \ref{fig:tsne} shows comparisons on the learned semantic feature distributions. CSAC obtains more discriminative and domain-agnostic information and generates class clear and domain compact semantic feature representations. We attribute it to the effectiveness of the collaborative semantic aggregation and calibration strategy
for domain invariance learning with distributed source domains. 

We then present insights on the proposed CSAC via showing the accuracy curves of the models on all the source datasets during training in Fig. \ref{fig:acc-train} (note that the target dataset is only used for testing the model performance). 
During semantic acquisition, each trained model is assigned to each separated domain for data distribution learning, and its accuracy on the learned domain improves rapidly (see the parts before model fusion in the subfigures (a-c)). 
The accuracy of the trained models assigned to Art, Cartoon, and Photo domain, on the target dataset, i.e., Sketch (red curve at dotted line), is 52.74\%, 68.47\%, 32.02\%, respectively, before model fusion. Then, the trained models are fused for semantic aggregation, each domain knowledge is fully gathered. The parameters of the fused model are then assigned to each trained model again for semantic calibration with local datasets. 
We unify semantic learning and alignment by repeating semantic aggregation and calibration alternately, the domain invariance from the separated domains is indirectly captured, making the accuracy of the fused model (subfigure (d)) on all the datasets improve gradually. By comparing the results before model fusion, the accuracy of the trained local model on the target dataset finally reaches improvement of more than 29\%, 13\%, and 50\%.

\section{Conclusions}\label{sec-con}
Training a generalizable model is a vital issue for the deep learning community. However, common practices of domain generalization rely on shared multi-source data, which may violate privacy policies in many real-world applications. This paper tackles the privacy-preserving problem of federated domain generalization, and presents a novel method for this challenging task with collaborative semantic aggregation and calibration. Our method unifies multi-source semantic learning and alignment in a collaborative way, distributed improving model generalization under careful privacy protection. In future, one may be demanded to collaboratively train a generalizable model by exploiting thousands of separated source datasets. Thus, our work sheds some light on this promising direction which lacks extensive research. It is important for many privacy-sensitive scenarios like finance and medical care. 
% The limitations of this work may be the homogeneous assumption of the data space, and future work may extend it to more complex scenarios.

\section*{Acknowledgments}
This work was supported in part by the National Key Research and Development Project (No. 2022YFC2504605), National Natural Science Foundation of China (62006207, U20A20387, 62037001), Young Elite Scientists Sponsorship Program by CAST (2021QNRC001), Zhejiang Provincial Natural Science Foundation of China (No. LZ22F020012), Major Technological Innovation Project of Hangzhou (No. 2022AIZD0147), Zhejiang Province Natural Science Foundation (LQ21F020020), Project by Shanghai AI Laboratory (P22KS00111), Program of Zhejiang Province Science and Technology (2022C01044), the StarryNight Science Fund of Zhejiang University Shanghai Institute for Advanced Study (SN-ZJU-SIAS-0010), and the Fundamental Research Funds for the Central Universities (226-2022-00142, 226-2022-00051).

\bibliographystyle{abbrv}
\bibliography{IEEEabrv.bib}

\end{document}